\documentclass[letterpaper, 10 pt, conference]{ieeeconf}  

\IEEEoverridecommandlockouts                             

\usepackage{array}
\usepackage{textcomp}
\usepackage{stfloats}
\usepackage{url}
\usepackage{verbatim}
\usepackage{graphicx}
\usepackage{cite}
\usepackage[utf8]{inputenc}
\usepackage[english]{babel}
\usepackage{setspace}
\usepackage{cite}
\usepackage{multirow}
\usepackage[table,xcdraw]{xcolor}
\usepackage{epsfig}
\usepackage{graphics}
\usepackage{graphicx}
\graphicspath{{fig/}}
\usepackage[mathscr]{euscript}
\let\labelindent\relax
\usepackage{enumitem}   
\usepackage{multicol}
\usepackage{multirow}
\usepackage{algorithm}
\usepackage{algorithmic}

\usepackage{amsthm} 
\usepackage{amsmath, amsfonts}
\usepackage{amssymb}
\usepackage[percent]{overpic}
\usepackage{siunitx}
\usepackage{soul}
\usepackage{sidecap}
\usepackage{caption}
\usepackage{subfigure}
\usepackage{float}

\usepackage{amsmath}
\usepackage{amssymb}
\usepackage{xparse}

\newcommand{\vect}[1]{\boldsymbol{#1}}
\newcommand{\mat}[1]{\boldsymbol{#1}}

\newcommand{\diffs}[3]{\frac{\partial^2 #1}{
		\ifx#2#3 
		\partial #2^2
		\else
		\partial #2 \partial #3
		\fi
}}

\newcommand{\grad}[2]{\boldsymbol{\nabla}_{#2}{#1}}

\newcommand{\hess}[2]{\boldsymbol{\nabla}_{#2}^{2}{#1}}

\newcommand{\R}{\mathbb{R}}

\newenvironment{carray}
{ \left( \begin{array}}
	{ \end{array} \right) }

\newcommand{\zerov}{\vect{0}}

\newcommand{\cv}{\vect{c}}

\newcommand{\fv}{\vect{f}}
\newcommand{\gv}{\vect{g}}

\newcommand{\hv}{\vect{h}}

\newcommand{\pv}{\vect{p}}

\newcommand{\qv}{{\vect{q}}}
\newcommand{\dqv}{\dot{\vect{q}}}
\newcommand{\ddqv}{\ddot{\vect{q}}}

\newcommand{\rv}{{\vect{r}}}

\newcommand{\sv}{\vect{s}}

\newcommand{\vv}{\vect{v}}

\newcommand{\xv}{\vect{x}}

\newcommand{\Fv}{\vect{F}}

\newcommand{\IIm}{\mat{I}}

\newcommand{\Am}{\mat{A}}

\newcommand{\Gm}{\mat{G}}

\newcommand{\Km}{\mat{K}}

\newcommand{\Mm}{\mat{M}}

\newcommand{\Rm}{\mat{R}}

\newcommand{\Tm}{\mat{T}}

\newcommand{\pvone}[1]{ \nabla_{#1}\, }
\newcommand{\pvtwo}[2]{ \nabla_{#2}\,#1 }
\newcommand{\parv}[2]{\expandafter\ifx\expandafter\relax
	\detokenize{#1}\relax\pvone{#2}\else\pvtwo{#1}{#2}\fi}

\NewDocumentCommand{\Rotm}{ O{} O{} O{} }{{}^{#2}\Rm_{#1}#3}
\NewDocumentCommand{\Transm}{ O{} O{} O{} }{{}^{#2}\Tm_{#1}#3}

\newcommand{\drm}{\mathrm{d}}

\newenvironment{sequation*}
    {\begin{equation*}\small
    }
    { 
    \end{equation*}
    }

\renewcommand{\baselinestretch}{.96} 

\title{\LARGE \bf
Nonlinear Modes as a Tool for Comparing the Mathematical Structure of Dynamic Models of Soft Robots 
}
\author{Pietro Pustina$^{1, 4}$, Davide Calzolari$^{2,3}$, Alin Albu-Schäffer$^{2,3}$, Alessandro De Luca$^{1}$, Cosimo Della Santina$^{3, 4}$ 
\thanks{
The reserach of Pietro Pustina was supported by the Sapienza SoftRobot project. The research of Alessandro De Luca was supported by the PNRR MUR project PE0000013-FAIR. The research of Cosimo Della Santina was supported by the Horizon Europe Program from Project EMERGE under Grant 101070918. The research of Davide Calzolari and Alin Albu-Schäffer was supported by the Advanced Grant M-Runners (ID: 835284) by the European Research Council.
$^{1}$Department of Computer, Control and Management Engineering, Sapienza University of Rome, Rome, Italy.
{\footnotesize pustina@diag.uniroma1.it; deluca@diag.uniroma1.it}
$^{2}$Institute of Robotics and Mechatronics, German Aerospace Center, 82234 Oberpfaffenhofen, Germany.
{\footnotesize davide.calzolari@dlr.de; alin.albu-schaeffer@dlr.de}
$^{3}$Department of Informatics, Technical University of Munich, 85748 Garching, Germany.
$^{4}$Department of Cognitive Robotics, Delft University of Technology, Delft, The Netherlands. 
{\footnotesize c.dellasantina@tudelft.nl}
}%
}

\begin{document}

\maketitle
\thispagestyle{empty}
\pagestyle{empty}

\begin{abstract}
Continuum soft robots are nonlinear mechanical systems with theoretically infinite degrees of freedom (DoFs) that exhibit complex behaviors. Achieving motor intelligence under dynamic conditions necessitates the development of control-oriented reduced-order models (ROMs), which employ as few DoFs as possible while still accurately capturing the core characteristics of the theoretically infinite-dimensional dynamics. However, there is no quantitative way to measure if the ROM of a soft robot has succeeded in this task. In other fields, like structural dynamics or flexible link robotics, linear normal modes are routinely used to this end. Yet, this theory is not applicable to soft robots due to their nonlinearities. 
In this work, we propose to use the recent nonlinear extension in modal theory --called eigenmanifolds-- as a means to evaluate control-oriented models for soft robots and compare them. To achieve this, we propose three similarity metrics relying on the projection of the nonlinear modes of the system into a task space of interest. We use this approach to compare quantitatively, for the first time, ROMs of increasing order generated under the piecewise constant curvature (PCC) hypothesis with a high-dimensional finite element (FE)-like model of a soft arm. Results show that by increasing the order of the discretization, the eigenmanifolds of the PCC model converge to those of the FE model. 
\end{abstract}
\section{Introduction}
Nonlinear modal analysis is a powerful tool to characterize the behavior of highly nonlinear mechanical systems~\cite{mikhlin2023nonlinear}, such as aerospace structures~\cite{kerschen2013nonlinear}, offshore towers~\cite{gavassoni2015nonlinear} and micro/nano-electromechanical systems~\cite{matheny2013nonlinear}. Despite the fact that its use has been historically for system analysis and design of structures undergoing small deformations, recent works~\cite{bjelonic2022experimental, coelho2022eigenmpc, wotte2023discovering} have used nonlinear modal theory in analyzing robots and in generating energy-efficient motions that asymptotically require no control effort to be sustained. 
%

\begin{figure}
    \centering
    \includegraphics[width=1\columnwidth, trim={0 0 0 0cm}, clip]{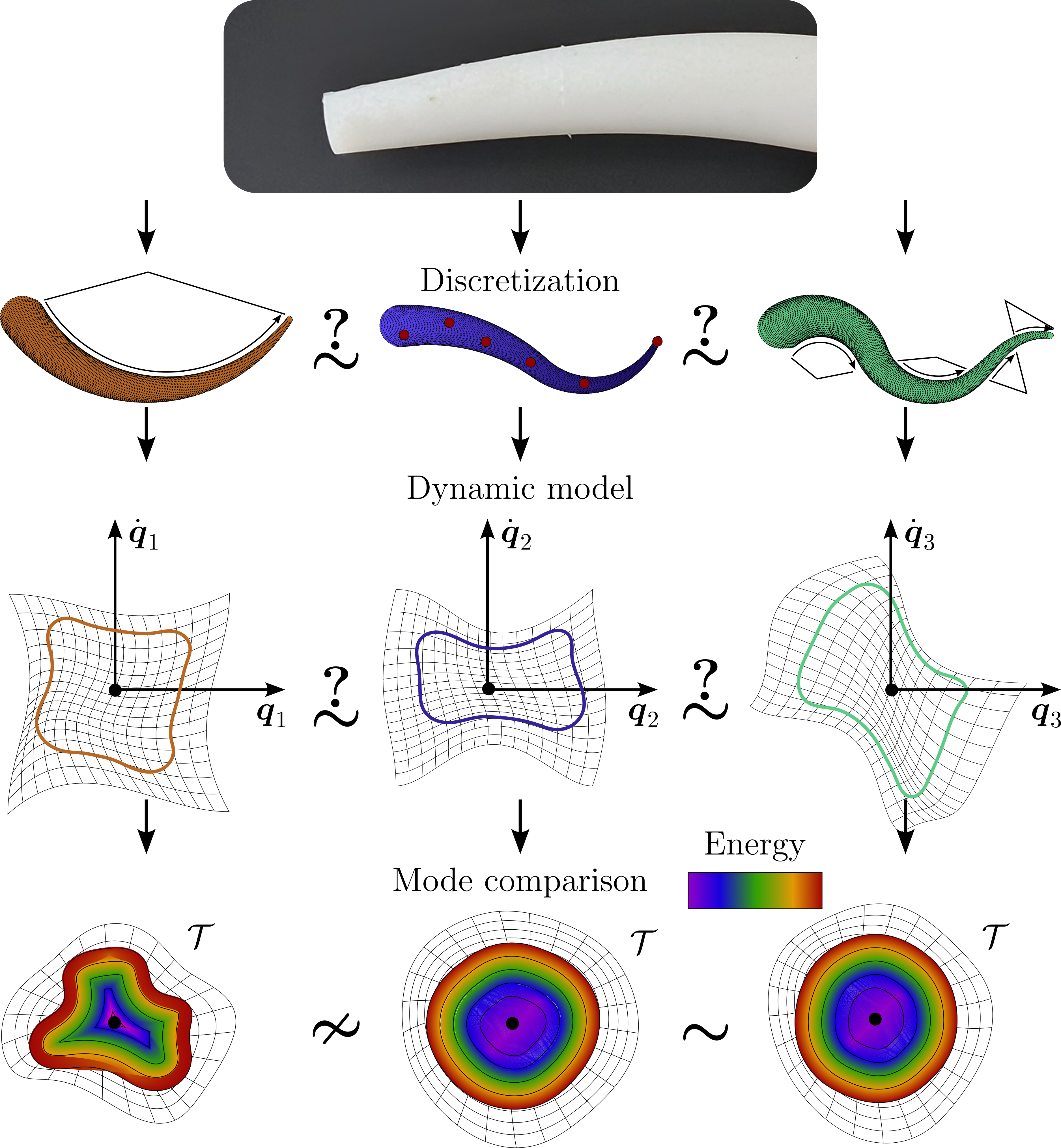}
    \caption{\small 
    Schematic of our approach using nonlinear modal analysis and eigenmanifold theory to compare different control-oriented models for the same soft robot. Models differ in assumptions and number of DoFs, making systematic comparison challenging. In the figure, $\sim$ indicates a similar structure, $\nsim$ indicates dissimilarity and a question mark denotes incomparability.
    }
    \label{fig:reduction scheme}
\end{figure}

Nonlinear modal analysis has found extensive application in flexible mechanical systems under the hypothesis of small deformations - especially in model reduction in structural mechanics have been studied~\cite{simpson2023use_collection, pesheck2002modal, touze2008reduced, kuether2014evaluating_collection, kuether2015evaluation}. In~\cite{touze2008reduced}, energy-frequency plots are used to compare ROMs of thin shells and plates. Finally, \cite{kuether2014evaluating_collection, kuether2015evaluation} use nonlinear modes to assess ROM convergence to FE models of specific structures. 

Continuum soft robots~\cite{della2021soft}, due to their high degree of deformability and inherent softness, are expected to exhibit a rich spectrum of modal behaviors.
However, none of these works are directly applicable to soft robots because they rely on nonlinear extensions of modal analysis built on assumptions that do not extend to soft robots and other multi-body systems.
%
%
Thus, the use of nonlinear modes in continuum soft robotics is still unexplored.

This work proposes the first application of nonlinear modal analysis to soft robotics by employing the recently developed concept of {\em eigenmanifolds} \cite{albu2020review} - which, contrary to classic theories, builds on hypotheses compatible with soft robotics. More specifically, with this work, we tackle the challenge of quantitatively comparing control-oriented reduced order models (ROMs) for continuum soft robots. In this way, it is also possible to compare control-oriented ROMs obtained from different discretization hypotheses, including those of~\cite{sadati2017control, grazioso2019geometrically, renda2020geometric, caasenbrood2023control}. The goal is to have a mathematical tool that can be used to quantify if two models have a similar mathematical structure. This is hard to do by inspection of the models because even their configuration spaces are going to be incompatible. Instead, we show in this work that modal theory can be used to this end, as illustrated in Figure~\ref{fig:reduction scheme}. 

We propose a two-stage strategy to achieve this objective. First, we operate a modal characterization of all the models to be compared. Then, we project the modes onto task space, where we introduce similarity measures to compare their shape and frequency content.
To summarize, the main contributions of the paper are as follows:
\begin{itemize}
    \item eigenmanifolds are proposed as a means of comparing control-oriented ROMs of soft robots derived from different discretization hypotheses;
    \item we propose a new energy-based continuation algorithm for eigenmanifold computation; 
    \item a systematic procedure is proposed for comparing eigenmanifolds of different ROMs, possibly obtained with different discretization techniques;
    \item we compute for the first time nonlinear modes of continuum soft robots.
\end{itemize}


\subsection{Notation}
Vectors and matrices are denoted in bold, while scalars are denoted in lowercase normal font. The $n \times n $ identity matrix is represented by $\IIm_{n}$. For a symmetric matrix $\Am = \Am^{T}$ with $\lambda_{\min}(\Am) > 0$, we write $\Am > 0$ for positive definiteness. To simplify the notation, arguments of functions are omitted when clear from the context. Furthermore, given a continuously differentiable function $f(\xv):  \mathbb{R}^{h} \rightarrow \mathbb{R}$, $\grad{f}{\xv} \in \mathbb{R}^{h}$ and $\hess{f}{\xv} \in \mathbb{R}^{h \times h}$ denote, respectively, the gradient and the Hessian of $f$.
We express the soft robot's dynamics as\footnote{As discussed in \cite{della2023model}, common and advanced modeling techniques used in soft robotics all yield equations of this form. Actuation and dissipation terms do not appear here because nonlinear modes are defined on the undamped and unactuated model. }
\begin{equation}\label{eq:equations of motion}
\Mm(\qv)\ddqv + \cv(\qv, \dqv) + \grad{V(\qv)}{\qv} = \zerov,
\end{equation}
where $\qv, \dqv, \ddqv \in \R^{n}$ denote the configuration variables and their time derivatives, $\Mm(\qv) \in \R^{n \times n}$ is the mass matrix and $\cv(\qv, \dqv) \in \R^{n}$ models Coriolis and centrifugal terms. Furthermore, $V(\qv)$ represents the potential energy, including, e.g., the effects of gravitational and elastic forces. We also denote the system energy as
\begin{equation}\label{eq:energy}
E(\qv, \dqv) = \frac{1}{2}\dqv^{T}\Mm(\qv)\dqv + V(\qv),
\end{equation}
and the solution of~\eqref{eq:equations of motion} at time $t$ from the initial condition $(\qv_{0}, \dqv_{0})$ as $\qv(t, \qv_{0}, \dqv_{0})$.

\section{Background: Nonlinear Modes}\label{sec:overview}
Nonlinear modal theory has been traditionally developed for mechanical systems with constant inertia matrix~\cite{kerschen2009nonlinear}, but it has recently been extended to multi-body mechanical systems~\cite{albu2020review}. Since continuum soft robots fall into the latter category, in the following, we briefly summarize the main concepts from~\cite{albu2020review}, which we refer the reader to for an exhaustive treatment of the topic.


Let $\qv_{eq}$ be a stable equilibrium configuration of~\eqref{eq:equations of motion}, i.e., 
\begin{equation*}
    \grad{V(\qv_{eq})}{\qv} = \zerov, \quad \Km(\qv_{eq}) = \hess{V(\qv_{eq})}{\qv} > 0,
\end{equation*}
which is equivalent to saying that $\qv_{eq}$ is a local minimizer of $E$ (for $\dqv = \zerov$). The linearization of~\eqref{eq:equations of motion} at $(\qv, \dqv) = (\qv_{eq}, \zerov)$ yields the second-order dynamics
\begin{equation}\label{eq:linearized eom}
    \Delta \ddqv + \Mm^{-1\!}(\qv_{eq})\Km(\qv_{eq})\Delta\qv = \zerov,
\end{equation}
where $\Delta \qv$ denotes a small increment of the configuration with respect to $\qv_{eq}$. 
Each eigenvector $\cv_i, \ i \in \{1, \cdots, n\}$ of $\Mm^{-1\!}(\qv_{eq})\Km(\qv_{eq})$ generates a local two-dimensional eigenspace 
\begin{equation*}
    ES_{i} = \mathrm{span}\left\{ \begin{carray}{c}
        \cv_{i}\\ \zerov        
    \end{carray}, \begin{carray}{c}
        \zerov\\ \cv_{i}        
    \end{carray}\right\}
\end{equation*}
with the property that when $(\Delta \qv, \Delta \dqv)$ is initialized in $ES_{i}$, the dynamics evolves according to the harmonic oscillator equations, namely
\begin{equation}\label{eq:harmonic oscillator}
    \Delta \ddqv + \omega_{i}^{2} \Delta \qv = \zerov,
\end{equation}
where $\omega_{i}^{2}$ is the eigenvalue associated with $\cv_{i}$. 
From the above equation, it can be shown that the solutions starting from $ES_{i}$ satisfy two important properties: (i) they are periodic with period $T_{i} = 2\pi/\omega_{i}$ and (ii) they are invariant, i.e., will remain in $ES_{i}$ at all times. These trajectories are called {\em linear modes} (LMs).

When moving to the nonlinear setting, each eigenspace $ES_{i}$ bends into a two-dimensional sub-manifold of the state space, called {\em eigenmanifold} and denoted as $\mathfrak{M}_{i}$. In analogy with the linear case, $\mathfrak{M}_{i}$ can be seen as a collection of invariant periodic trajectories of~\eqref{eq:equations of motion}, called hereinafter nonlinear modes (NMs). However, NMs can have different periods, while all LMs have the same period.
For the purpose of this paper, we need a further notion, namely that of eigenmanifold generator.
Given $\mathfrak{M}_{i}$, its generator is defined as 
\begin{equation*}
    \mathcal{G}_{\mathfrak{M}_{i}} = \left\{ (\qv, \dqv) \in \mathfrak{M}_{i} \,\,|\,\, \dqv = \zerov \right\}.
\end{equation*}
In a nutshell, $\mathcal{G}_{\mathfrak{M}_{i}}$ is the subset of $\mathfrak{M}_{i}$ containing all points with zero velocity. Each element of $\mathcal{G}_{\mathfrak{M}_{i}}$ is an initial rest configuration associated with a periodic motion that remains in $\mathfrak{M}_{i}$ at all times. The elements of $\mathcal{G}_{\mathfrak{M}_{i}}$ can be seen as the nonlinear counterparts of the eigenvector $\cv_{i}$. Differently from the linear case, where we can generate $ES_{i}$ with only one vector, the nonlinear case needs an entire set of points, i.e., $\mathcal{G}_{\mathfrak{M}_{i}}$, to generate $\mathfrak{M}_{i}$. An important property of $\mathcal{G}_{\mathfrak{M}_{i}}$ is that it can be parameterized by the energy, implying that every NM can itself be parameterized by its energy. To simplify the notation, we will henceforth drop the subscript $i$ and denote the NM at $E$ as $\mathcal{G}_{\mathfrak{M}}(E)$.
\section{Eigenmanifold computation}\label{sec:manifold computation}
For the computation of NMs, analytical and numerical approaches have been proposed in the literature, see~\cite{kerschen2009nonlinear} for a thorough review on possible strategies. 

Remarkably, following a two-stage procedure, we can easily extend numerical approaches used for the computation of NMs also for that of eigenmanifolds. First, we compute the generator of $\mathfrak{M}$. Then, we take each point of $\mathcal{G}_\mathfrak{M}$ as an initial condition to simulate the dynamics forward in time and compute the corresponding NM, thus obtaining a slice of $\mathfrak{M}$.

In this work, for the computation of $\mathcal{G}_{\mathfrak{M}}$ we use a modified version of the numerical continuation algorithm presented in~\cite{peeters2009nonlinear}. Superscripts are used to denote the iteration index.
Each generator is computed using a predictor-corrector algorithm that starts from the direction given by the eigenvector $\cv_{i}$ of the linearized dynamics at the equilibrium. 
During the first iteration only, the linearized dynamics~\eqref{eq:linearized eom} is used to compute a point close to $\mathcal{G}_{\mathfrak{M}}$ and $\qv_{eq}$. In particular, an initial configuration increment $\Delta \qv_{0}$ 
is computed such that the linearized energy takes a small incremental step.
In this way, the trajectory of~\eqref{eq:equations of motion} from $\qv_{0}^{1} = \qv_{eq} + \Delta \qv_{0}$ with zero velocity stays close to $ES$ and has approximately the same period $T^1$ of that of $\Delta \qv_{0}$. Then, $(T^{1}, \qv_{0}^{1}, \zerov)$ is refined with a Newton-Raphson iteration scheme that seeks a solution of the periodicity constraint
\begin{equation}\label{eq:periodicity constraint}
    \rv(\qv_{0}, T) = \begin{carray}{c}
        \qv_{0} - \qv(T, \qv_{0}, \zerov)\\[2pt]
        - \dqv(T, \qv_{0}, \zerov)
    \end{carray} = \zerov,
\end{equation}
in the unknowns $(\qv_{0}, \zerov) \in \mathcal{G}_{\mathfrak{M}}$ and $T \in (0, \infty)$. At step $j \geq 2$, a prediction for the current solution $(T^{j}, \qv_{0}^{j},\zerov)$ is computed along the direction given by the tangent vector to the previous solution $(T^{j-1},\qv_{0}^{j-1},\zerov)$. The prediction is then corrected with a shooting procedure. In~\cite{peeters2009nonlinear}, the shooting procedure aims at solving $\rv(\qv_{0}^{j}, T^{j}) = \zerov$, which is an over-constrained system of $2n$ equations in $n+1$ unknowns. Since the predictor step requires fairly small energy increments to guarantee convergence of the procedure, the computation of the generator can be quite time consuming. To speed up the evaluation, we propose to constrain the energy increment between consecutive iterations. In particular, during the correction stage we solve the augmented system
\begin{equation}\label{eq:predictor equation}
    \begin{carray}{c}
        \rv(\qv_{0}^{j}, T^{j})\\[2pt]
        E(\qv_{0}^{j}, \zerov) - E(\qv_{0}^{j-1}, \zerov) - \Delta E
    \end{carray} = \zerov,
\end{equation}
being $\Delta E>0$ a desired energy step. This simple modification has proven to significantly reduce the computation time, without affecting stability of the procedure. 
Since convergence of the scheme is guaranteed only in the vicinity of $\mathcal{G}_{\mathfrak{M}}$, $\Delta E$ must be properly selected to prevent numerical instability. To this end, a simple adaptive rule has been adopted. In particular, $\Delta E$ is initialized to a reference value $\Delta \Bar{E}$ and a maximum number $N_{\max}$ of iterations is defined for the correction. If convergence is not achieved within $N_{\max}$ iterations, $\Delta E$ is halved and the correction repeated. 
\section{Metrics}\label{sec:metrics}
In this work, we aim at comparing how the eigenmanifolds of a continuum soft robot change with the order and type of the discretization adopted and thus with the number DoFs. A direct comparison in configuration space is not possible in general, unless the discretizations have the same number of DoFs. 
\begin{figure}
    \centering
    \includegraphics[width=1\columnwidth]{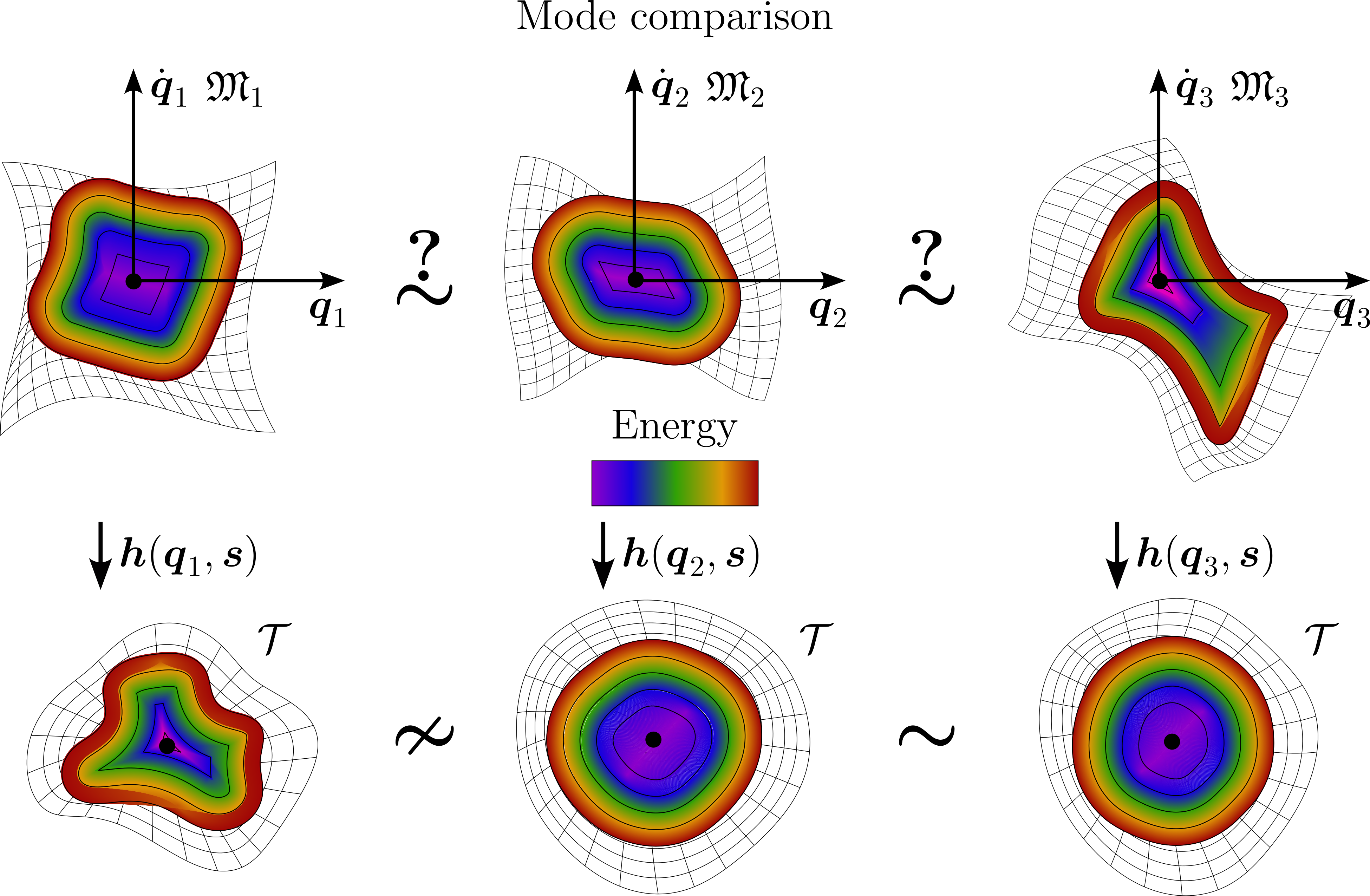}
    \caption{\small Schematic representation of the eigenmanifold projection. Given three control-oriented ROMs, their eigenmanifolds $\mathfrak{M}_{1}$, $\mathfrak{M}_{2}$ and $\mathfrak{M}_{3}$ cannot be directly compared in configuration space. To overcome this issue, we introduce a constant dimensional task space $\hv(\qv, \sv)$ in which $\mathfrak{M}_{1}$, $\mathfrak{M}_{2}$ and $\mathfrak{M}_{3}$ are projected and here define similarity measures.}
    \label{fig:metrics scheme}
\end{figure}
Even though two models have the same number of DoFs, their configuration spaces may not be directly comparable. For example, consider the dynamic model of a soft robot obtained using a piecewise constant curvature and an affine approximation of the strain. Even if the discretization order is chosen such that the models have the same number of DoFs, the configuration variables will still belong to different spaces, making a direct comparison impossible. 

We propose to overcome this issue by introducing a task space function $\hv(\qv, \sv) : \R^{n} \times \mathcal{S} \rightarrow \mathcal{T}$, where $\sv \in \mathcal{S} \subseteq \R^{q}$ denotes a set of hyper-parameters for $\mathcal{T} \subseteq \R^{m}$ and $m$ has always the same dimension, independently of the dimension $n$ of $\qv$. Given $\hv(\qv, \sv)$, any eigenmanifold can be projected into $\mathcal{T}$, resulting in a collection of periodic orbits. Because all trajectories of an eigenmanifold are periodic by definition, their projection into $\mathcal{T}$ is a periodic trajectory in $\mathcal{T}$ with same period. Therefore, the representation of $\mathfrak{M}$ into $\mathcal{T}$ is a sub-manifold of $\mathcal{T}$ with properties similar to $\mathfrak{M}$. The approach is illustrated in Fig.~\ref{fig:metrics scheme}.

For example, consider the direct kinematics of a slender soft arm. Regardless of the discretization technique used, it must always be possible to reconstruct the robot backbone (using a single hyper-parameter $s$). In other words, we know the direct kinematics mapping 
\begin{equation}\label{eq:direct kinematics slender robot}
    \hv(\qv, s) = \pv(\qv, s) = \begin{carray}{c}
        p_{x}(\qv, s)\\
        p_{y}(\qv, s)\\
        p_{z}(\qv, s)\\
    \end{carray}:\R^{n} \times \R \rightarrow \R^{3},
\end{equation}
where $\pv(\qv, s)$ is the position of the point at distance $s \in [0, L]$ from the base, being $L$ the robot  length at rest.

Because all the level sets of the energy 
\begin{equation*}
    L_{\Bar{E}} = \left\{ (\qv, \dqv) \in \mathfrak{M} \,\,|\,\, E(\qv, \dqv) = \Bar{E} \right\}
\end{equation*}
give raise to evolutions that trace the same path in configuration space and differ only in a phase shift, a single point of $L_{\Bar{E}}$ should be projected. The generator points are good candidates for this purpose because they are all characterized by an initial zero velocity, which allows to get rid of any phase shift due to the initialization in $\mathfrak{M}$.

Given the task manifold, it is now possible to define similarity measures between two eigenmanifolds $\mathfrak{M}_{1}$ and $\mathfrak{M}_{2}$. These can depend on different properties that one is interested in comparing, such as their shape and spectral content. A common measure to compare two curves $A(t)$ and $B(t)$ belonging to a metric space $S$ is the Fréchet distance~\cite{tao2021comparative},
\begin{equation}\label{eq:frechet distance}
    \phi(A, B) = \inf_{\alpha, \beta} \max_{u \in [0, 1]} \left\{ d(A(\alpha(u)), B(\beta(u))) \right\},
\end{equation}
where $d(\cdot, \cdot)$ is the distance function of $S$. In a nutshell, the above metric looks at all possible reparameterizations of $A$ and $B$ and takes the one that minimizes their maximum distance. Given $\hv(\qv, \sv)$ and~\eqref{eq:frechet distance}, we can define the following measures
\begin{equation}\label{eq:modal distance}
    \fv(E, \sv) = \phi(\hv(\mathcal{G}_{\mathfrak{M}_{1}}(E), \sv), \hv(\mathcal{G}_{\mathfrak{M}_{2}}(E), \sv)),
\end{equation}
and
\begin{equation}\label{eq:modal integral distance}
    \Fv(E) = \int_{\mathcal{S}} \fv(E, \vv) \drm \vv,
\end{equation}
where the Fréchet distance is applied componentwise to the rows of $\hv(\qv, \sv)$. In the following, we refer to $\fv(E, \sv)$ and $\Fv(E)$, respectively, as the modal Fréchet distance and modal integral Fréchet distance.
The measure $\fv$ allows us to compare two modes for a given value of $E$ and $\sv$. On the contrary, $\Fv$ averages this measure over the entire space of hyper-parameters for $\hv(\qv, \sv)$. It is also worth remarking that other choices instead of $\phi$ could have been made, see for example~\cite{tao2021comparative}. 

Consider again the case of a slender soft arm and take $\hv(\qv, s) = \pv(\qv, s)$. Suppose we are interested in quantifying how close two eigenmanifolds are. A first estimate can be obtained by looking at the tip evolution only, i.e., $\fv(E, L)$. However, $\fv(E, L)$ provides only a partial indication of their similarity, because $\fv(E, L) = \zerov$ does not imply $\fv(E, s) = 0$ for any other point $s \neq L$. To obtain a more complete measure of similarity, we introduce $\Fv(E)$. Note that if $\Fv(E) = \zerov$, then $\fv(E, \sv) = \zerov$ for all $\sv \in \mathcal{S}$, which implies that the NMs produce the same motion for all the system particles.

To compare the spectra, we can adopt a similar strategy. To this end, we consider the minimum value of the magnitude-squared coherence defined as follows
\begin{equation}
    \gamma(A, B) = \min_{f}\frac{| g_{AB}(f) |^{2}}{g_{AA(f)}g_{BB(f)}},
\end{equation}
where $g_{AB}$ is the Cross-spectral power density between $A$ and $B$, and $g_{AA}$ and $g_{BB}$ the auto power spectral density of $A$ and $B$, respectively. At a given frequency, the coherence measures the extent to which $B$ can be estimated from $A$ through a linear process. Note that unlike $\phi(A, B)$, $\gamma(A, B)$ is not a distance function, as it is not symmetric and lower values indicate less correlation between the two signals, thus worst similarity. By applying $\gamma$ to the components of $\hv$ we define the modal coherence as
\begin{equation}
    \gv(E, \sv) = \gamma(\hv(\mathcal{G}_{\mathfrak{M}_{1}}(E), \sv), \hv(\mathcal{G}_{\mathfrak{M}_{2}}(E), \sv)),
\end{equation}
and the modal integral coherence
\begin{equation}
    \Gm(E) = \int_{\mathcal{S}} \gv(E, \vv) \drm \vv.
\end{equation}
Considerations analogous to those for $\fv$ and $\Fv$ hold. 

The integral measures defined above can also be integrated over the energy domain to obtain an overall measure of the similarity between the eigenmanifolds, i.e.,
\begin{equation*}
    \Fv_{E} = \int_{E_{\qv_{eq}}}^{E_{\max}} \Fv(E) \drm E,
\end{equation*}
and
\begin{equation*}
    \Gm_{E} = \int_{E_{\qv_{eq}}}^{E_{\max}} \Gm(E) \drm E,
\end{equation*}
where $E_{\max}$ is the maximum energy level considered during analysis. In future works, we will explore the use of these measure as loss terms in a learning algorithm for the computation of control-oriented ROMs.

\section{Simulation Results}
\label{sec:simulations}

In this section, we apply the above metrics to compute and compare the eigenmanifolds of a slender continuum soft robot. The goal is to investigate the differences between PCC ROMs obtained by increasing the resolution of the strain and a FE rigid model having a large number of DoFs. For the PCC models, we enhance the resolution by increasing the number of bodies per unit of length at rest. The robot has a cylindrical shape with radius $r = 0.02~[\si{\meter}]$, uniform mass density $\rho = 1062~[\si{\kilogram\per\cubic\meter}]$, and rest length $L = 0.4~[\si{\meter}]$. Furthermore, we consider a linear elastic material with Young modulus $Y = 0.66~[\si{\mega\pascal}]$ and Poisson ratio $\nu = 0.5$. The arm is mounted with the base rotated with the gravitational field so that the straight stress-free configuration is a stable equilibrium, and motion occurs in the $(x,z)$ plane.

All the bodies of the PCC models share the same physical properties, except for the rest length which is chosen as $L/n$, being $n$ the discretization order. The maximum number of bodies we consider is $n_{\max} = 5$. For each of these models, we compute all their $n \leq n_{\max}$ eigenmanifolds, using the procedure outlined in Section~\ref{sec:manifold computation}. The rigid robot consists of ten cylindrical bodies, each one with length $L/10$ and mass $m = \displaystyle\rho\pi r^2 (L/10)~[\si{\kilogram}]$. Elasticity is introduced by assuming linear elastic revolute joints with uniform stiffness coefficients $k = Y \pi r^4 10/(4L)~[\si{\newton\meter\per\radian}]$. 
Because the PCC models have at most five bodies, we compute only the first five eigenmanifolds of the rigid model, ordered in ascending order by oscillation frequency. The eigenmanifold computation is perfomed with a desired energy step of $\Delta E = 0.05~\si{[\joule]}$ until the energy reaches the maximum value $E_{\max} = 1~[\si{\joule}]$. In the following, we use the direct kinematics of the backbone~\eqref{eq:direct kinematics slender robot} as task space function. The integral measures $F(E)$ and $G(E)$ are approximated through a Gaussian quadrature rule with ten Gaussian points.

\begin{figure*}[!ht]
    \centering
    \subfigure[{}]{
        \includegraphics[height=20pt, trim={0 9cm 0 0}, clip]{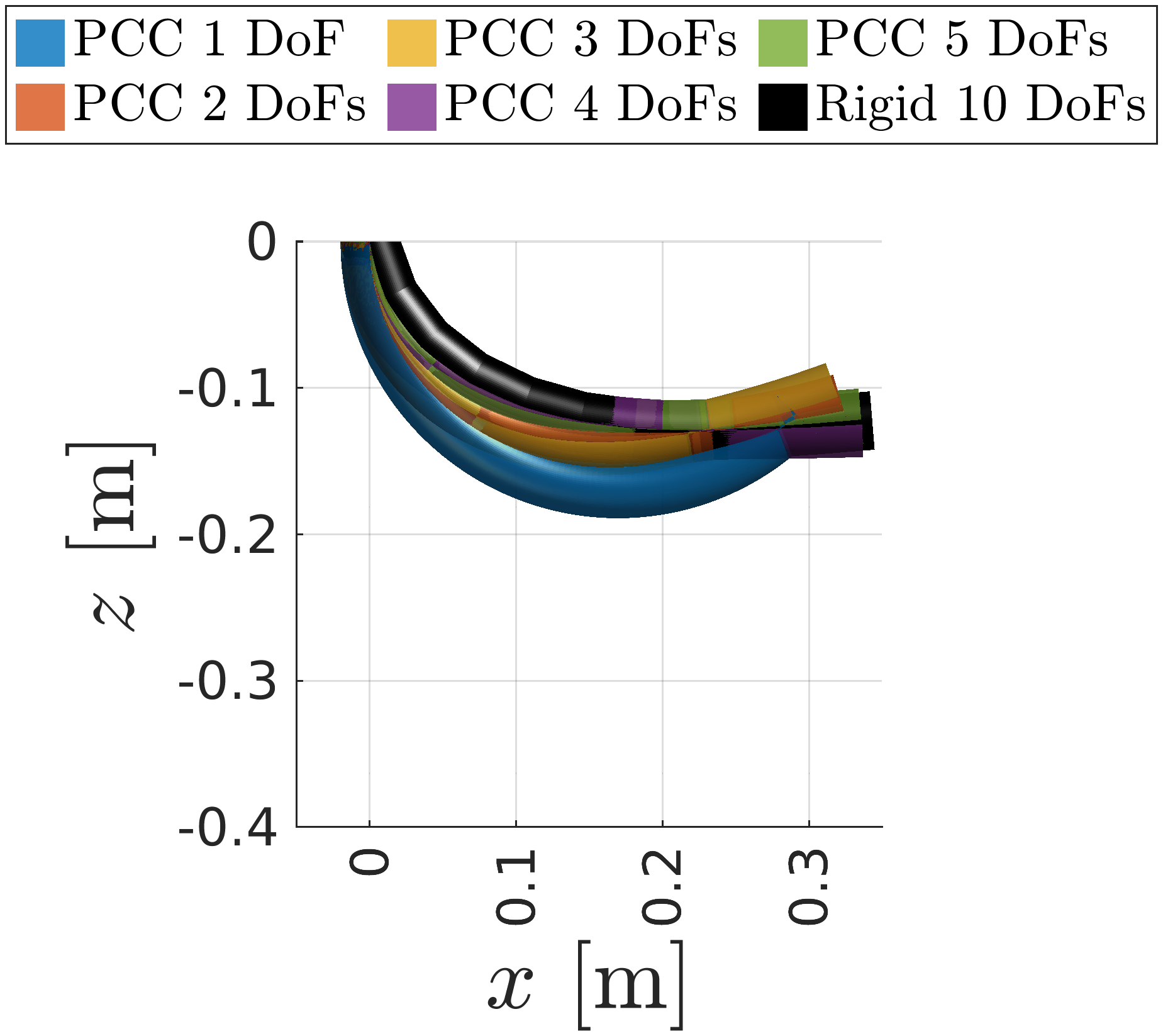}
    }\\\setcounter{subfigure}{0}\vspace{-1cm}
    \subfigure[{Mode 1; $t = 0T_{1}$}]{
        \includegraphics[height = 95px, trim={0, 0, 2.5cm, 1.5cm}, clip]{matlab/ModeSolver/fig/results_strobo_mode_1_time_1.png}
        \label{fig:mode 1:time 1}
    }\hspace{-1cm}\hfill
    \subfigure[{$t = 0.2 T_{1}$}]{
        \includegraphics[height = 86px, trim={0cm, 0, 0cm, 0cm}, clip]{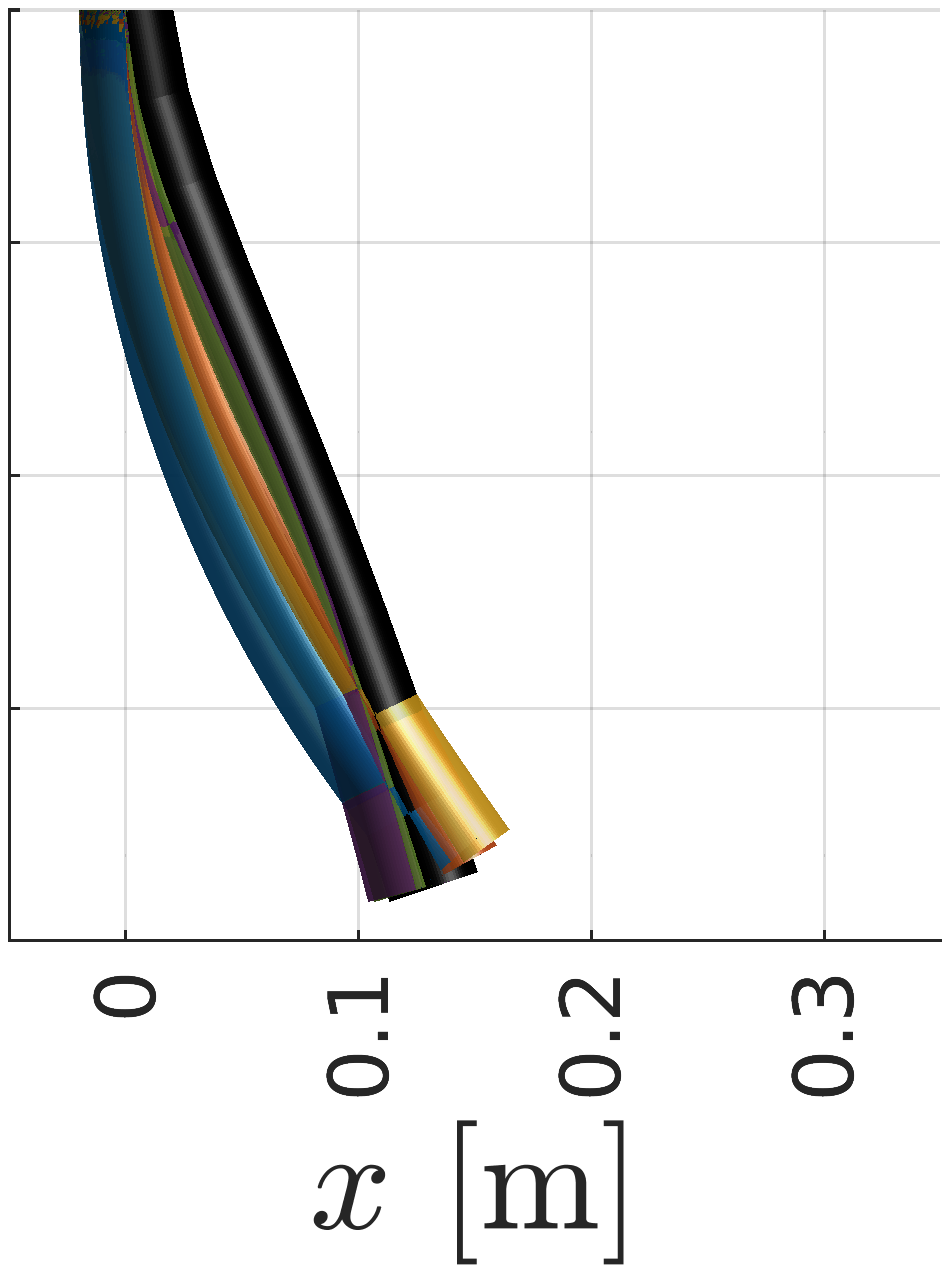}
        \label{fig:mode 1:time 2}
    }\hspace{-1cm}\hfill
    \subfigure[{$t = 0.3 T_{1}$}]{
        \includegraphics[height = 86px, trim={0cm, 0, 0cm, 0cm}, clip]{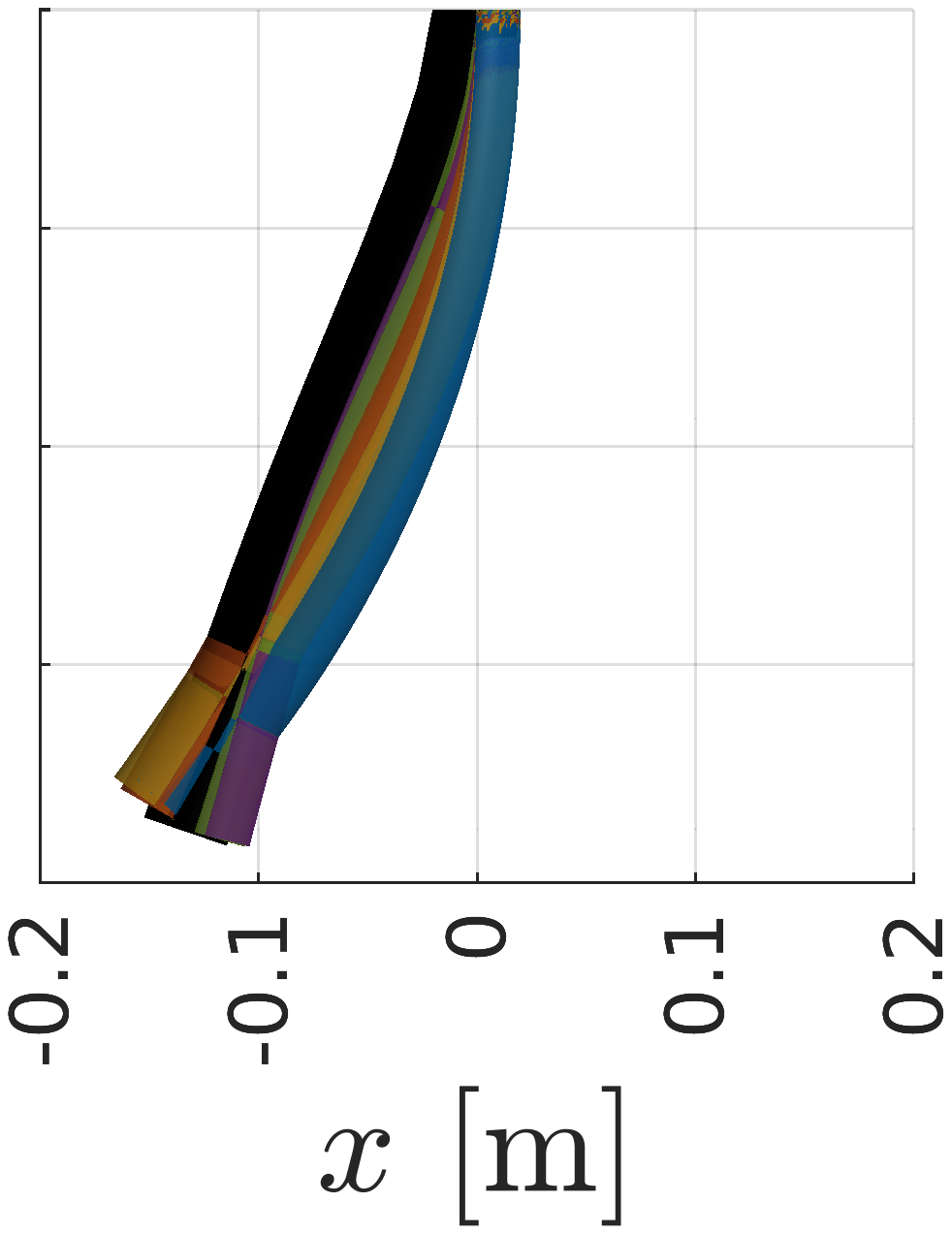}
        \label{fig:mode 1:time 3}
    }\hspace{-1cm}\hfill
    \subfigure[{$t = 0.5 T_{1}$}]{
        \includegraphics[height = 86px, trim={0cm, 0, 0cm, 0cm}, clip]{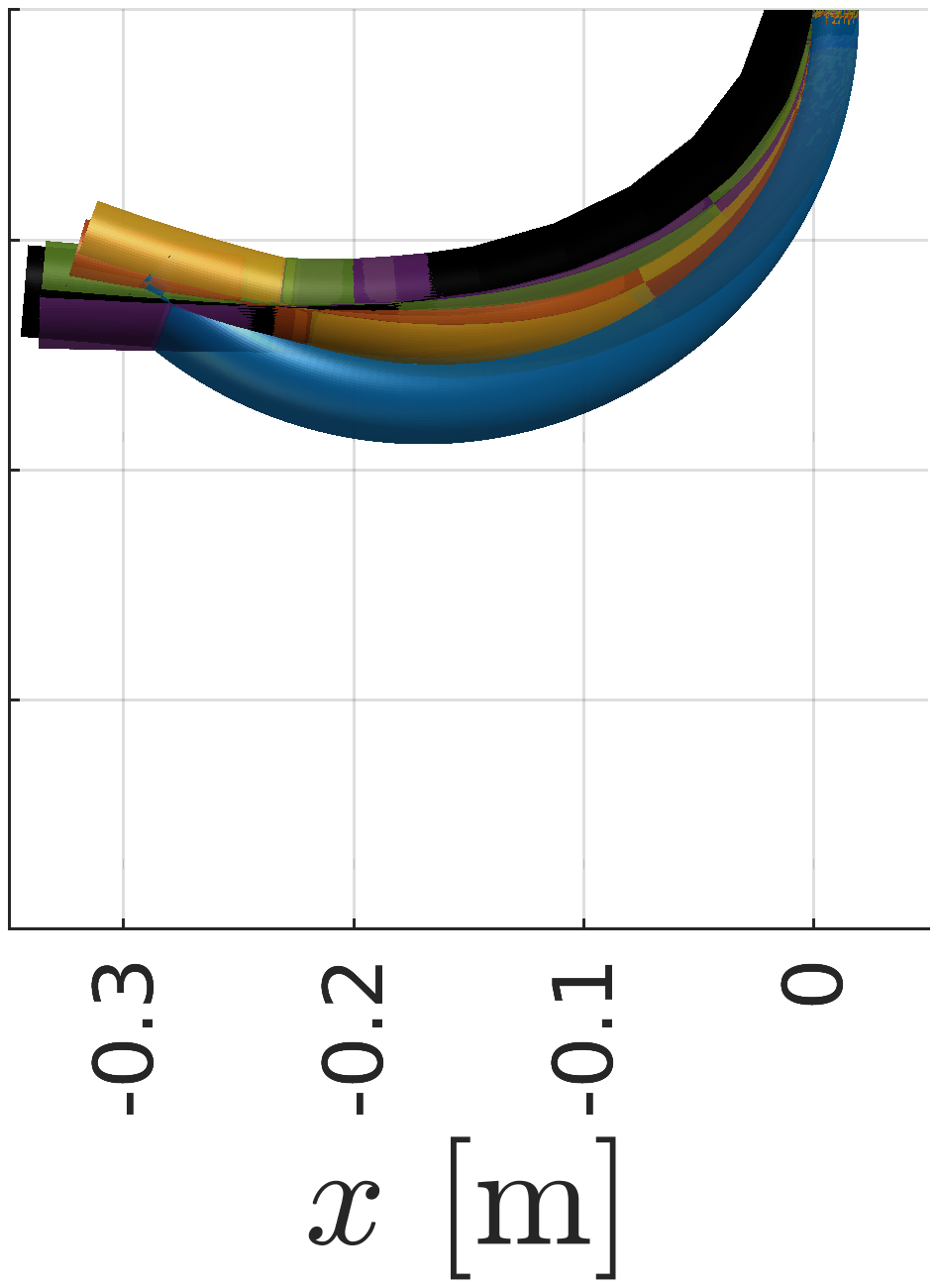}
        \label{fig:mode 1:time 4}
    }\hspace{-1cm}\hfill
    \subfigure[{$t = 0.7 T_{1}$}]{
        \includegraphics[height = 86px, trim={0cm, 0, 0cm, 0cm}, clip]{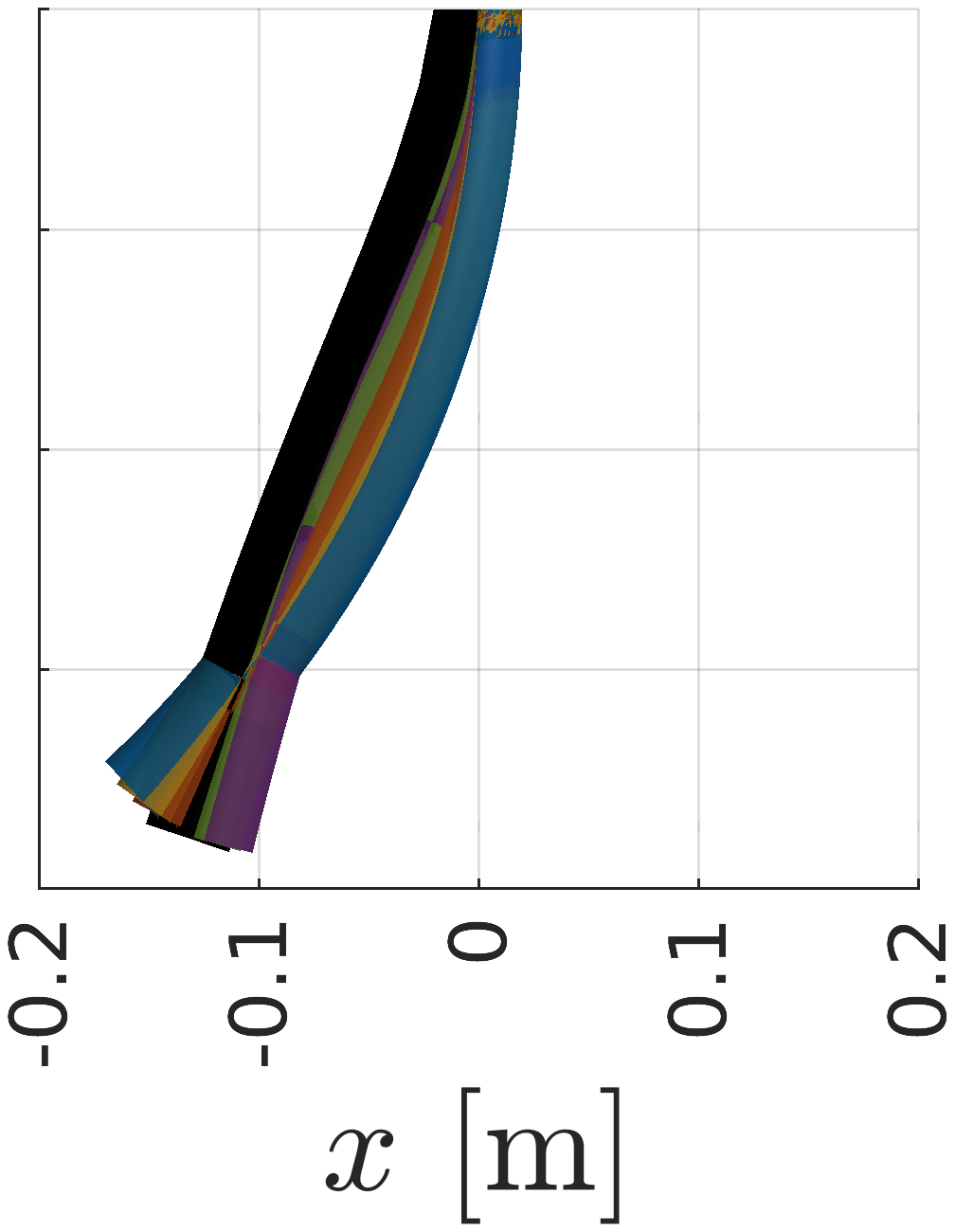}
        \label{fig:mode 1:time 5}
    }\hspace{-1cm}\hfill
    \subfigure[{$t = 0.8 T_{1}$}]{
        \includegraphics[height = 86px, trim={0cm, 0, 0cm, 0cm}, clip]{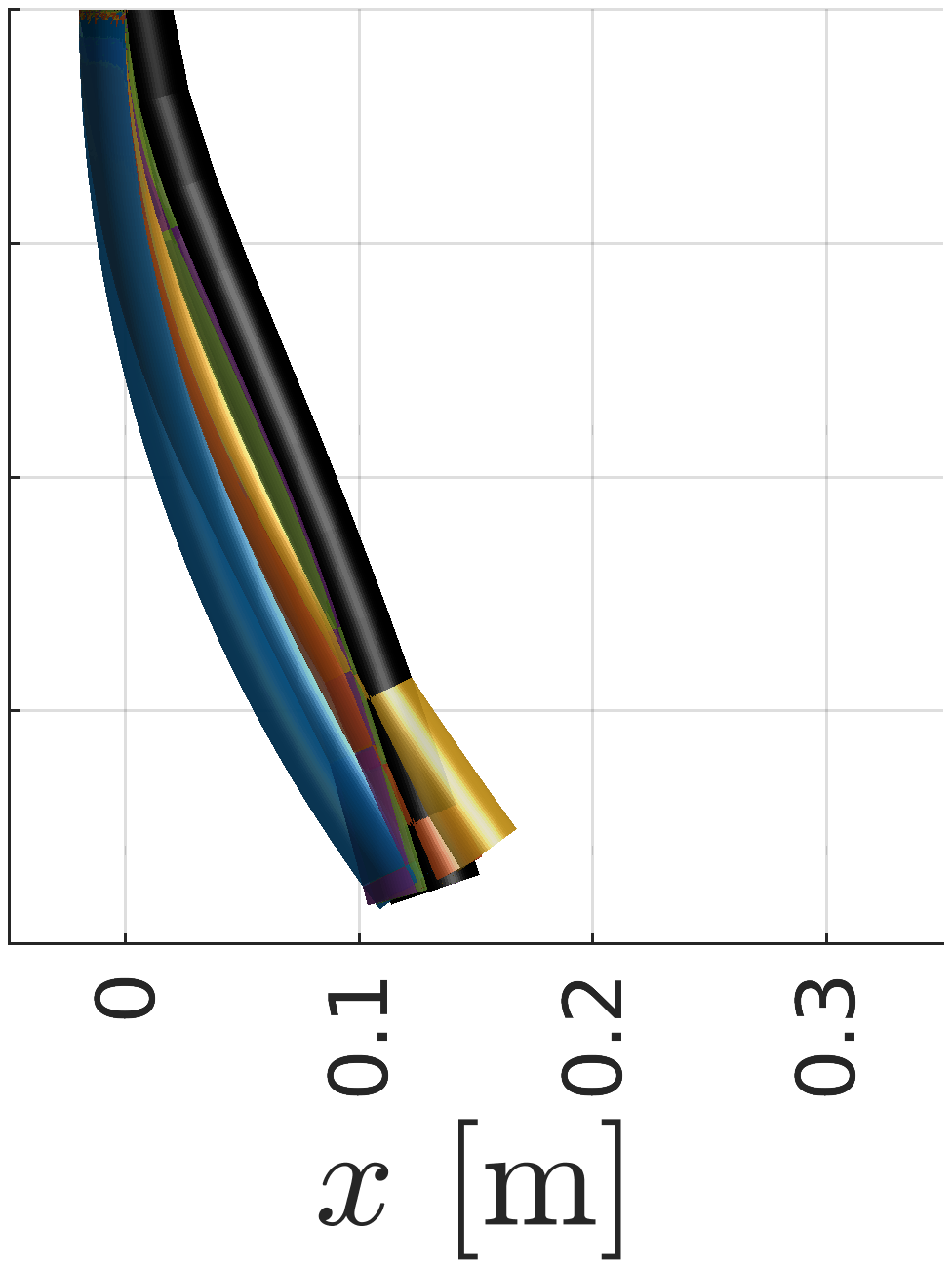}
        \label{fig:mode 1:time 6}
    }\hspace{-1cm}\hfill
    \subfigure[{$t = 0.9 T_{1}$}]{
        \includegraphics[height = 86px, trim={0cm, 0, 0cm, 0cm}, clip]{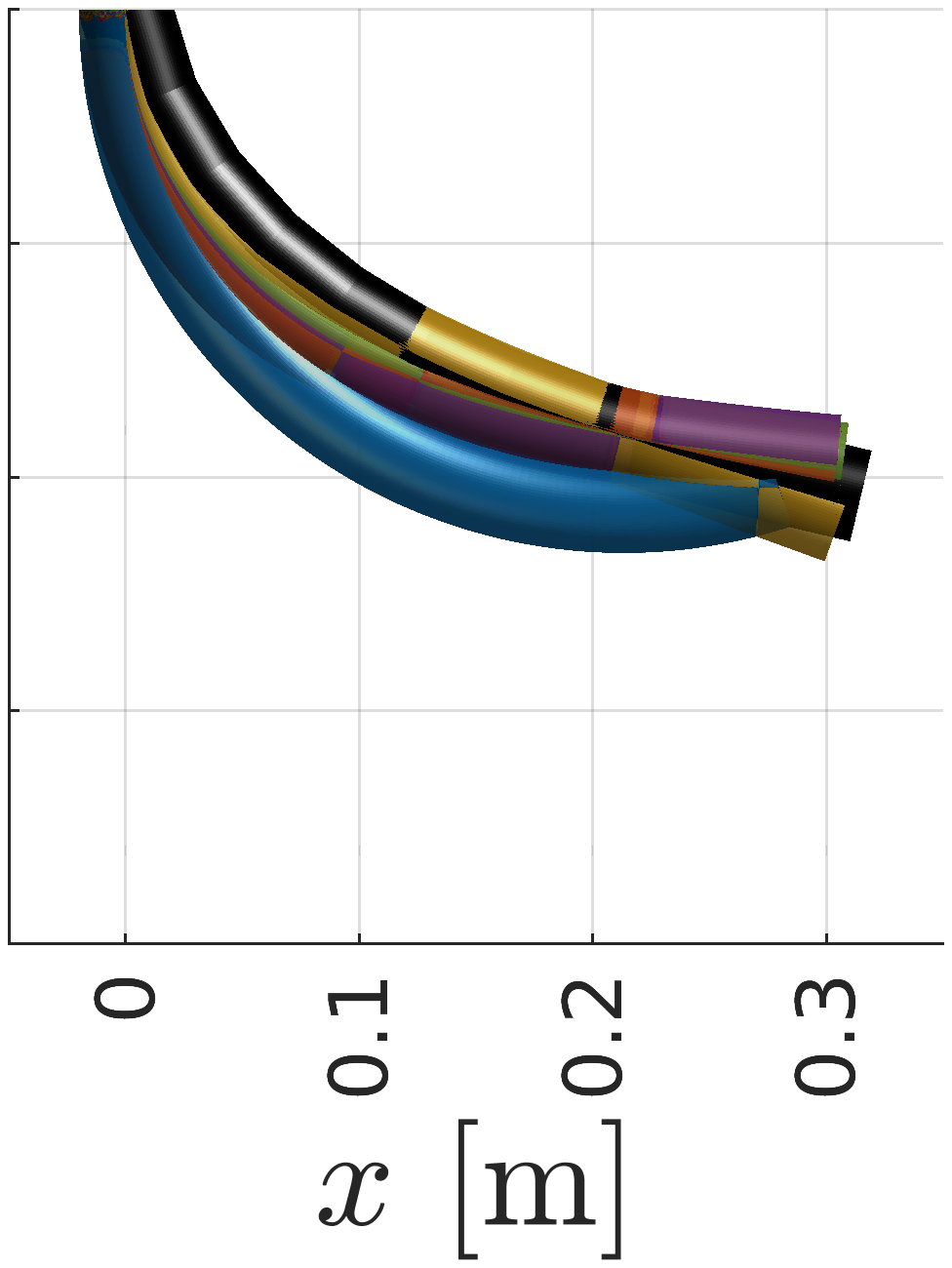}
        \label{fig:mode 1:time 7}
    }\\\vspace{-0.3cm}
    \subfigure[{Mode 2; $t = 0T_{2}$}]{
        \includegraphics[height = 95px, trim={0, 0, 2.5cm, 1.5cm}, clip]{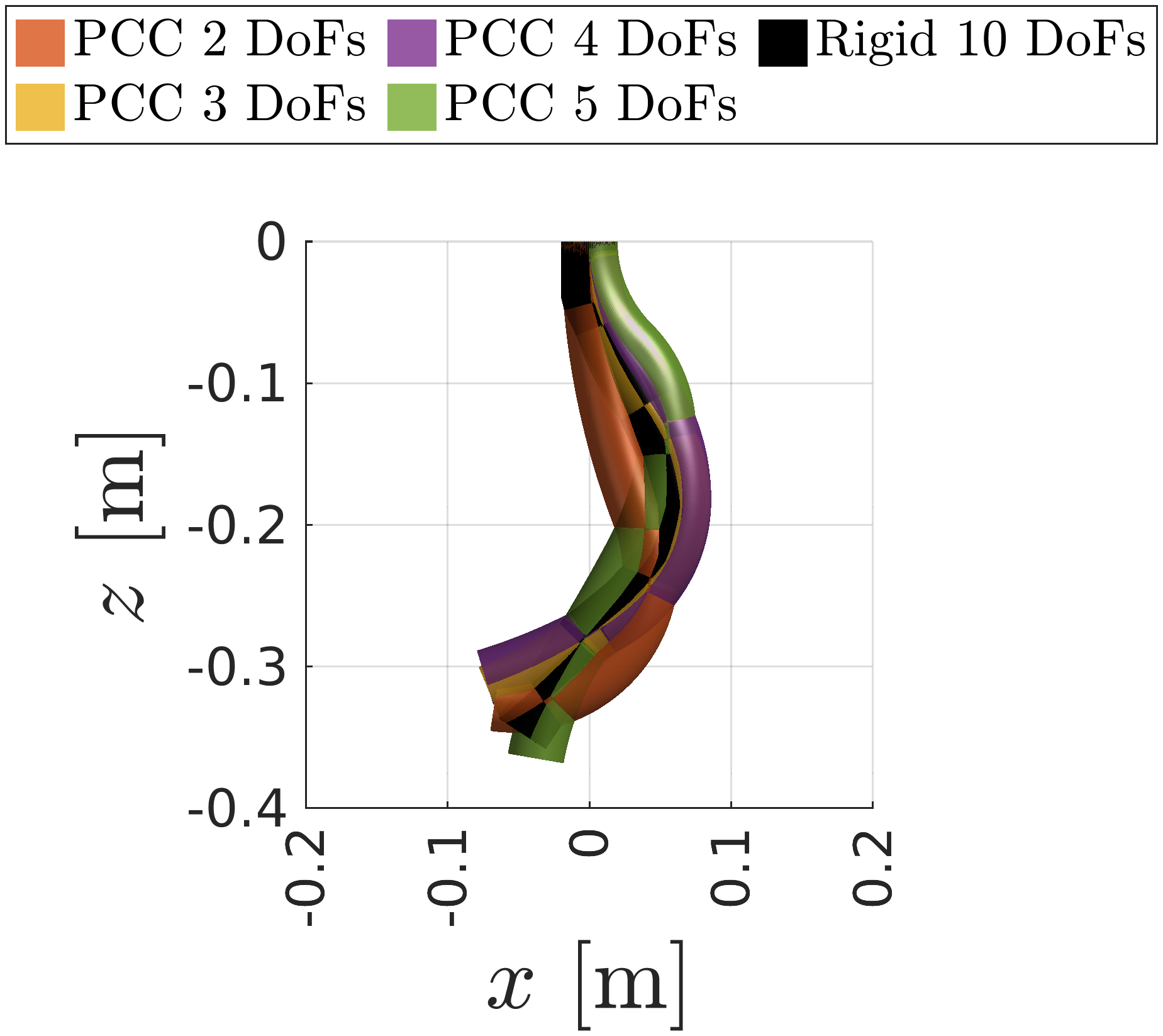}
        \label{fig:mode 2:time 1}
    }\hspace{-1cm}\hfill
    \subfigure[{$t = 0.2 T_{2}$}]{
        \includegraphics[height = 86px, trim={0cm, 0, 0cm, 0cm}, clip]{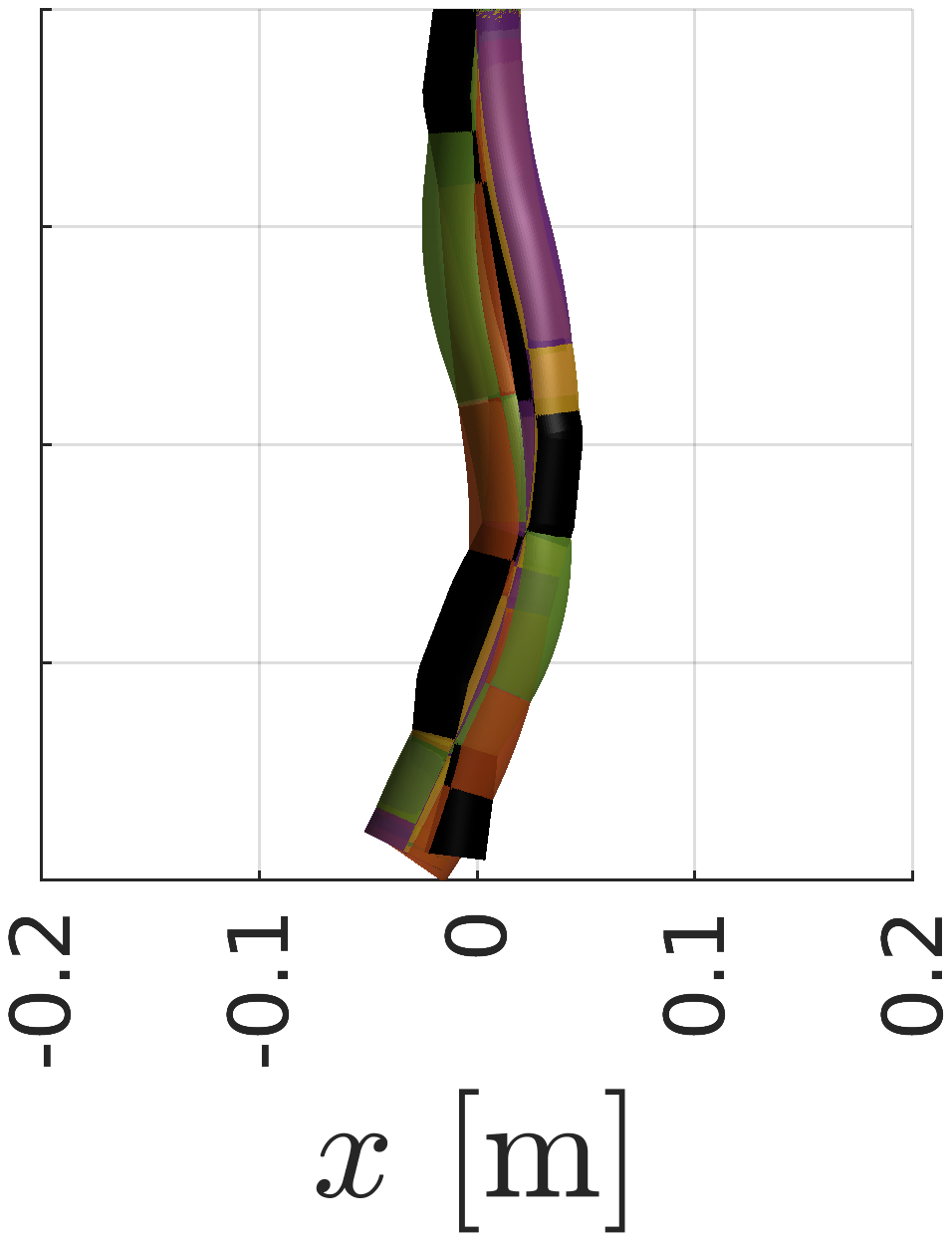}
        \label{fig:mode 2:time 2}
    }\hspace{-1cm}\hfill
    \subfigure[{$t = 0.3 T_{2}$}]{
        \includegraphics[height = 86px, trim={0cm, 0, 0cm, 0cm}, clip]{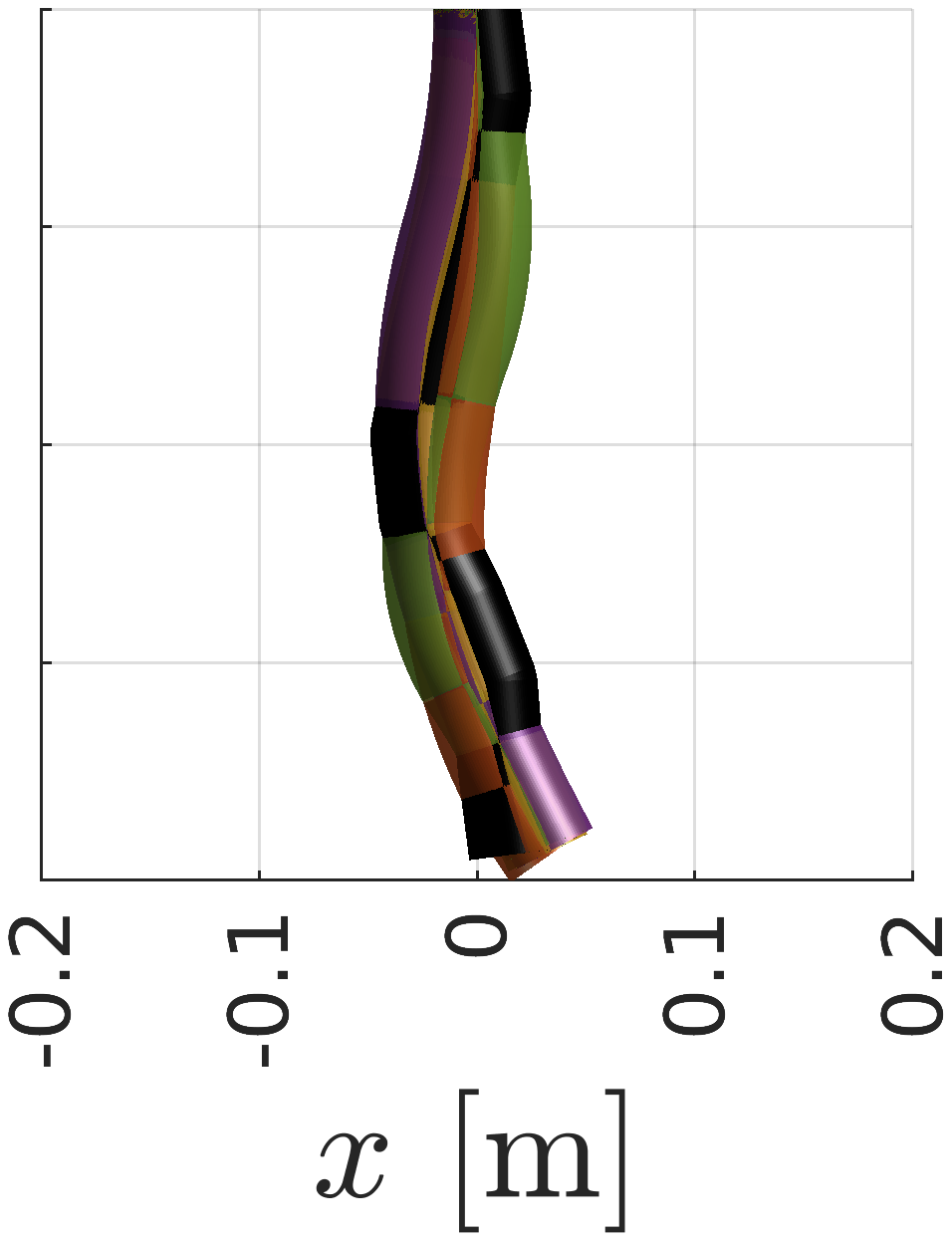}
        \label{fig:mode 2:time 3}
    }\hspace{-1cm}\hfill
    \subfigure[{$t = 0.5 T_{2}$}]{
        \includegraphics[height = 86px, trim={0cm, 0, 0cm, 0cm}, clip]{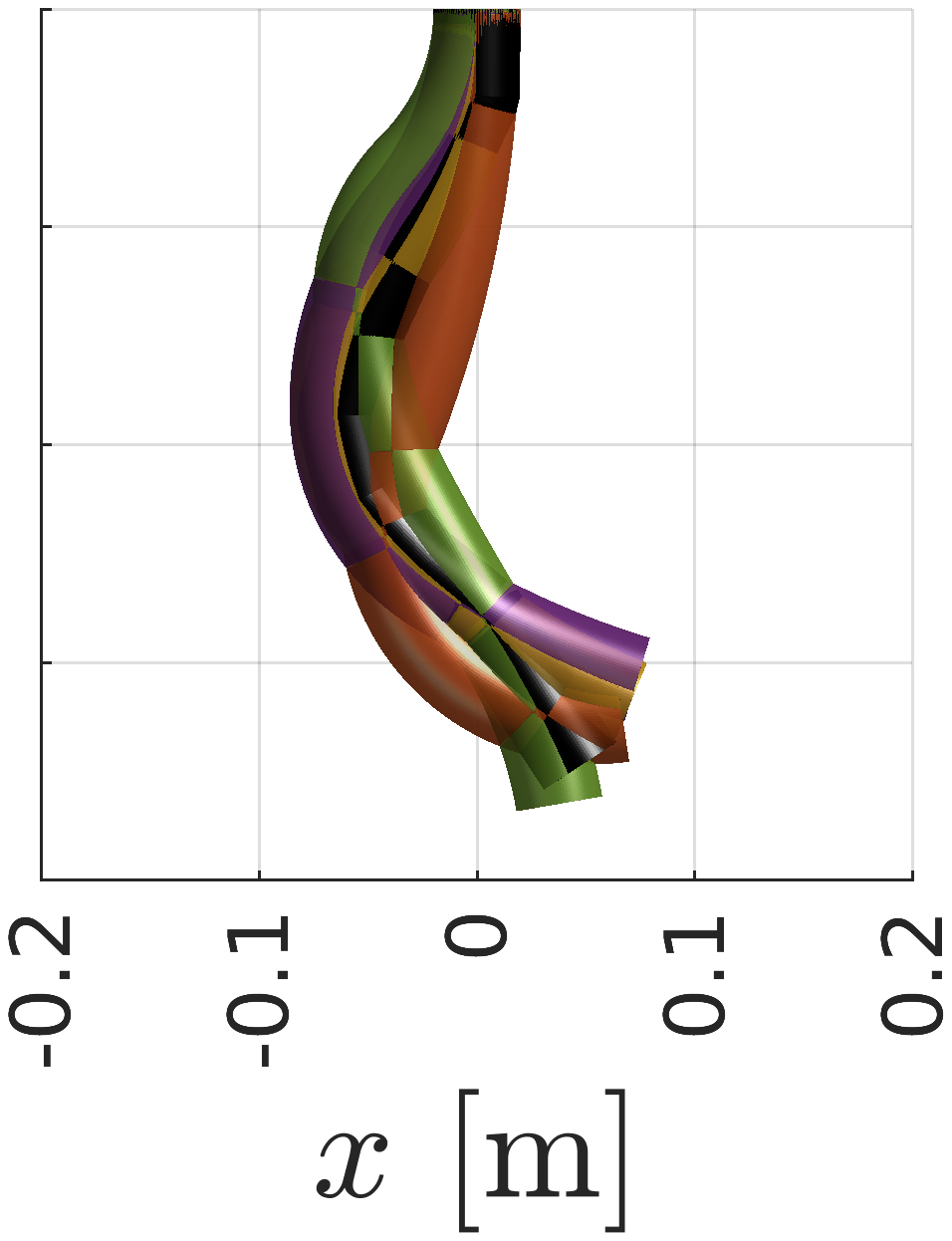}
        \label{fig:mode 2:time 4}
    }\hspace{-1cm}\hfill
    \subfigure[{$t = 0.7 T_{2}$}]{
        \includegraphics[height = 86px, trim={0cm, 0, 0cm, 0cm}, clip]{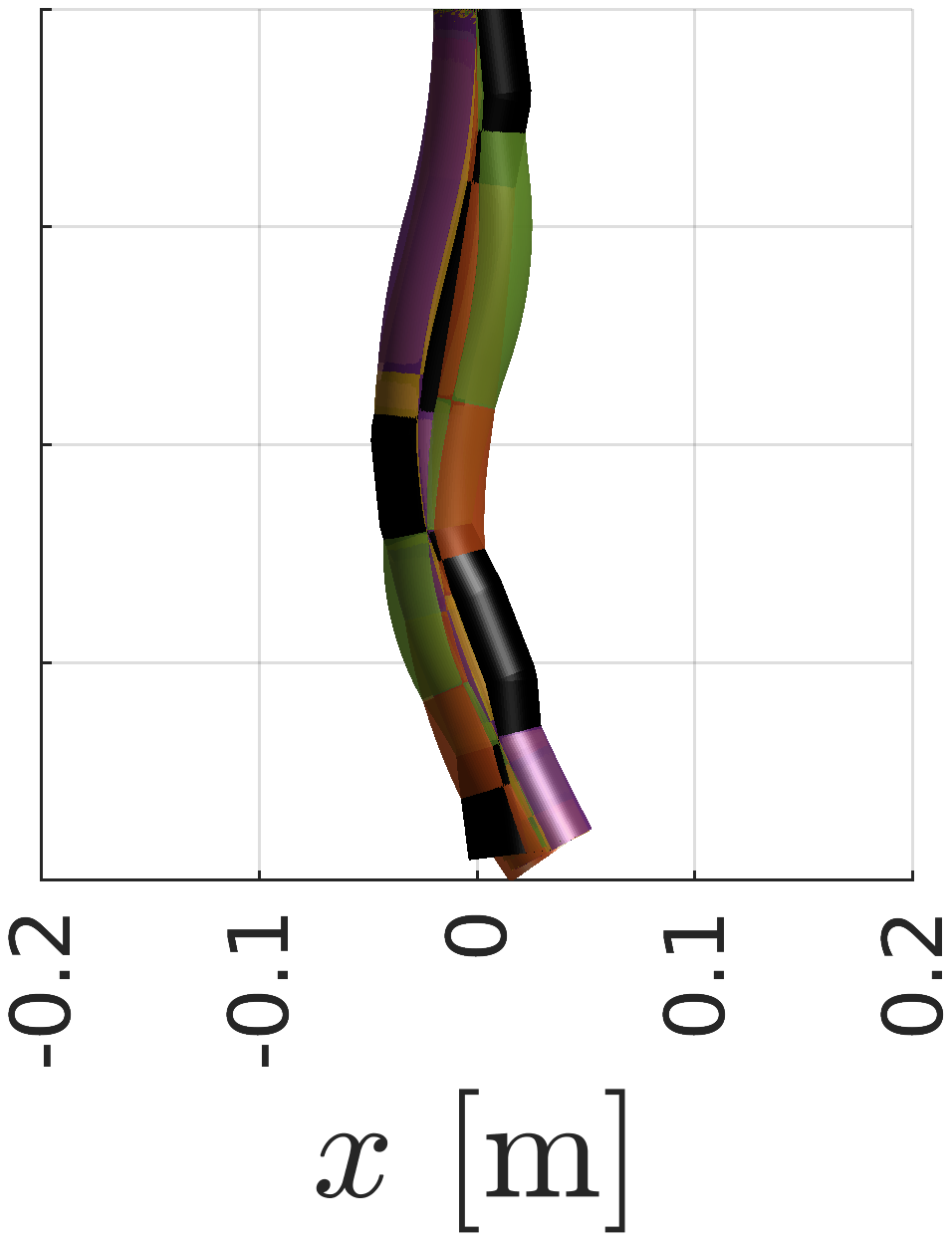}
        \label{fig:mode 2:time 5}
    }\hspace{-1cm}\hfill
    \subfigure[{$t = 0.8 T_{2}$}]{
        \includegraphics[height = 86px, trim={0cm, 0, 0cm, 0cm}, clip]{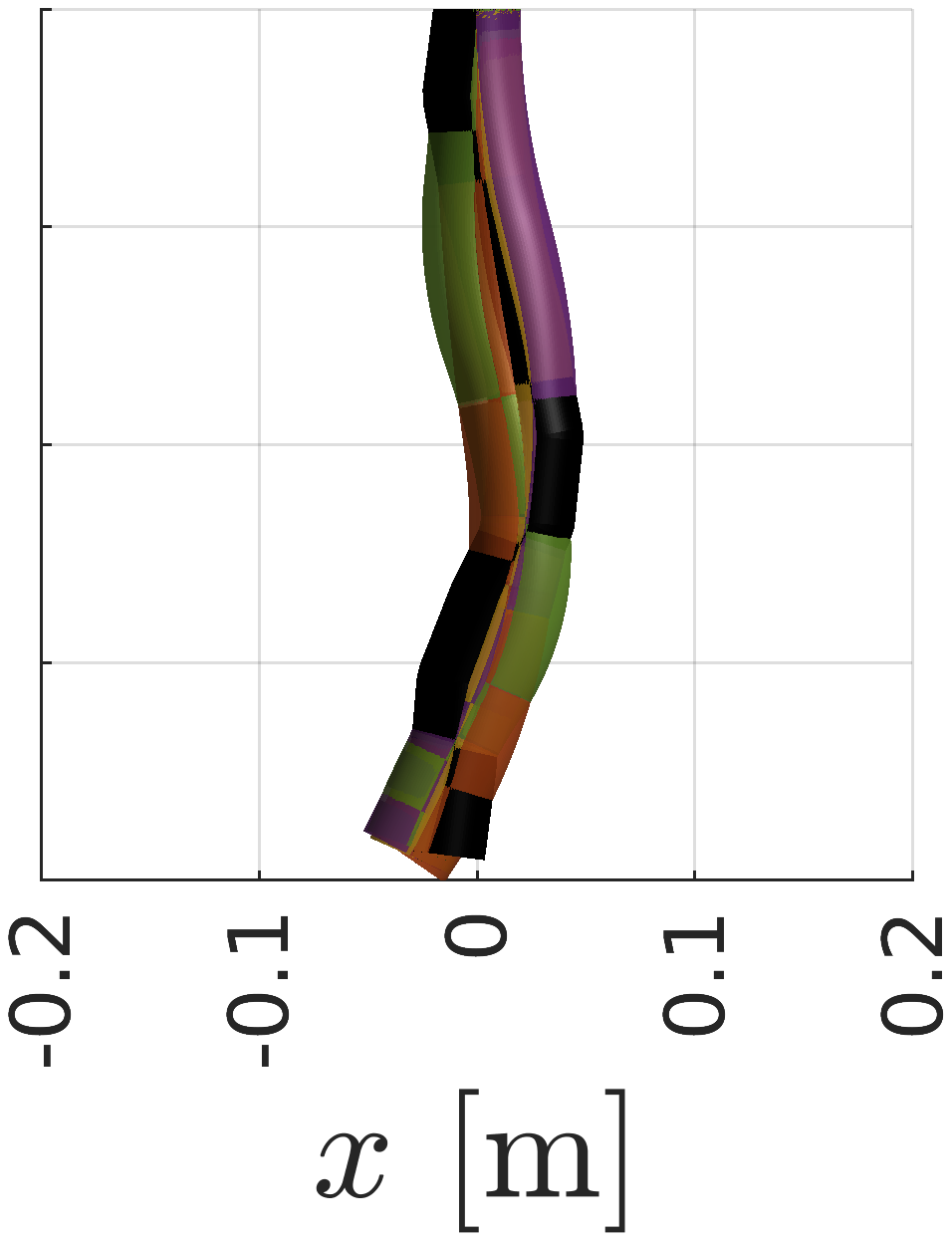}
        \label{fig:mode 2:time 6}
    }\hspace{-1cm}\hfill
    \subfigure[{$t = 0.9 T_{2}$}]{
        \includegraphics[height = 86px, trim={0cm, 0, 0cm, 0cm}, clip]{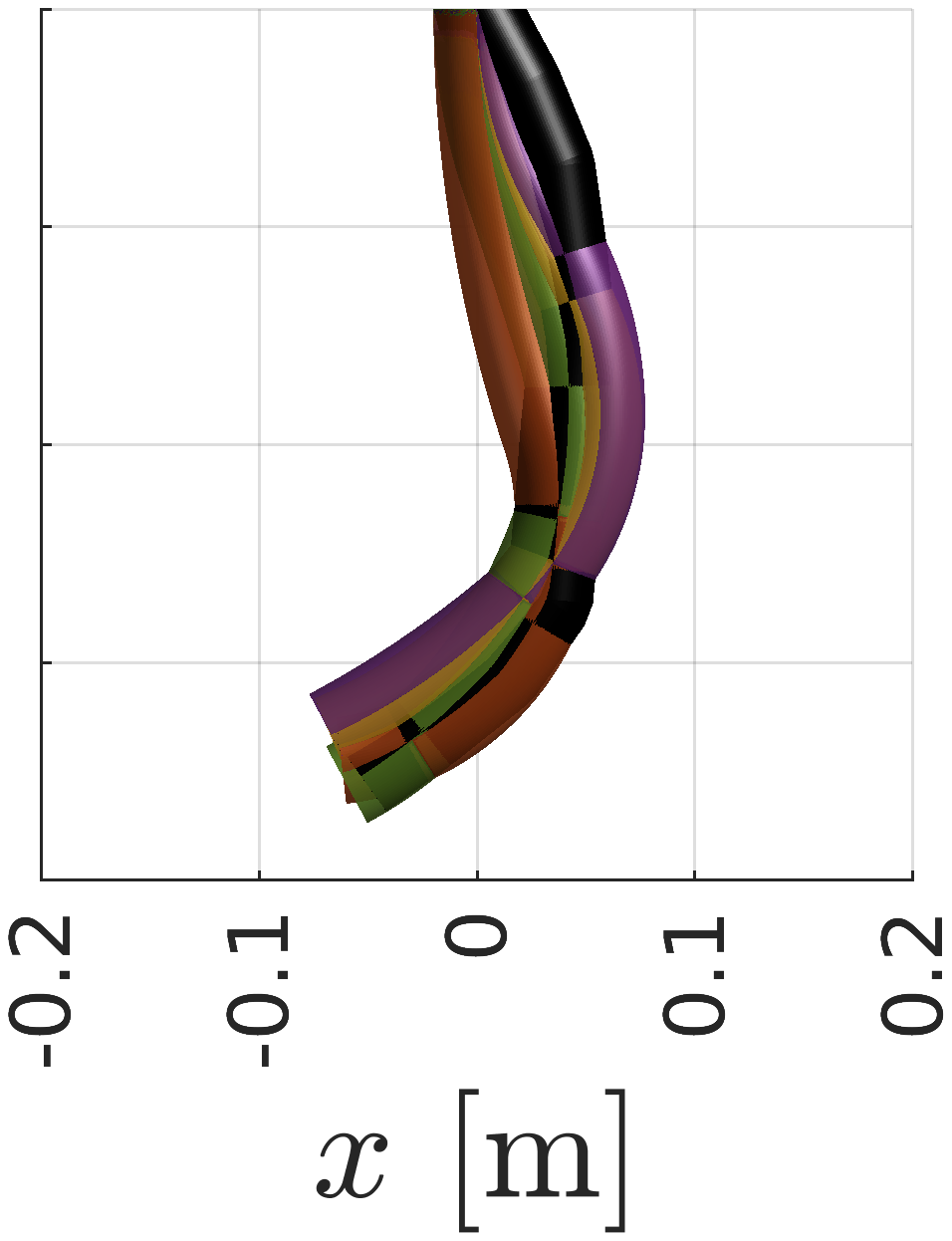}
        \label{fig:mode 2:time 7}
    }\vspace{-0.2cm}\\\hspace{0.2cm}
    \subfigure[{Mode 3; $t = 0T_{3}$}]{
        \includegraphics[height = 95px, trim={0, 0, 0.9cm, 1.5cm}, clip]{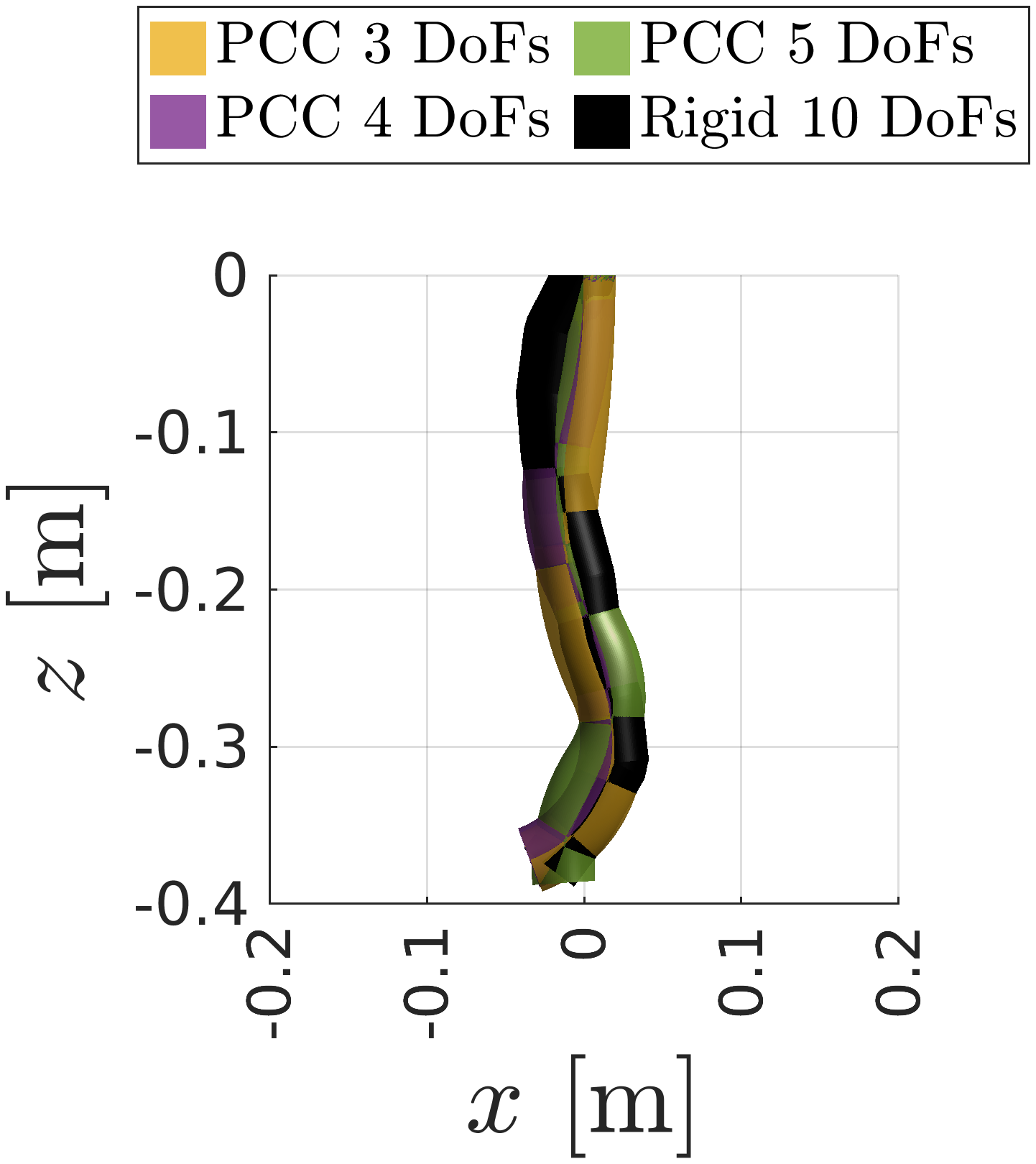}
        \label{fig:mode 3:time 1}
    }\hspace{-1cm}\hfill
    \subfigure[{$t = 0.2 T_{3}$}]{
        \includegraphics[height = 86px, trim={0cm, 0, 0cm, 0cm}, clip]{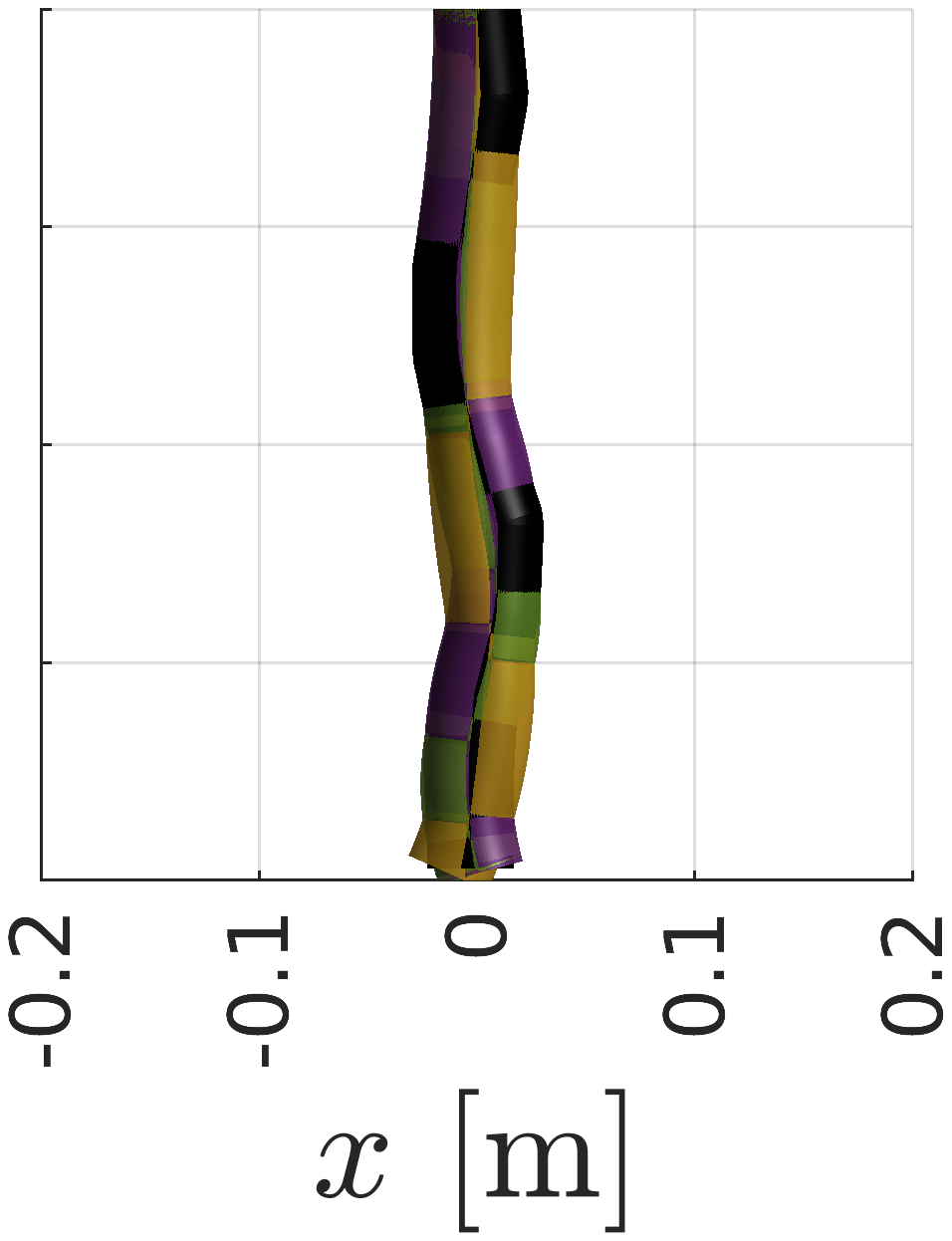}
        \label{fig:mode 3:time 2}
    }\hspace{-1cm}\hfill
    \subfigure[{$t = 0.3 T_{3}$}]{
        \includegraphics[height = 86px, trim={0cm, 0, 0cm, 0cm}, clip]{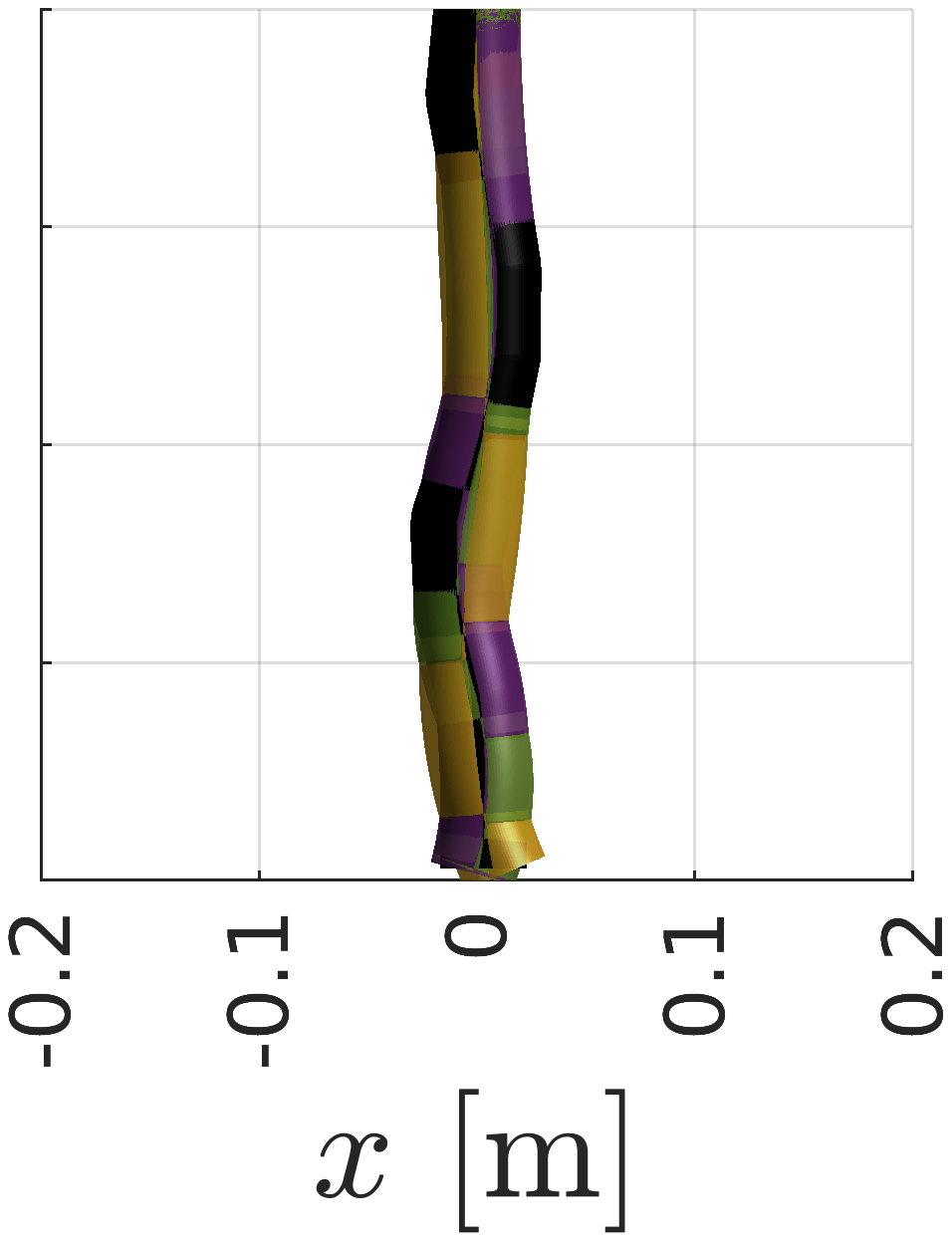}
        \label{fig:mode 3:time 3}
    }\hspace{-1cm}\hfill
    \subfigure[{$t = 0.5 T_{3}$}]{
        \includegraphics[height = 86px, trim={0cm, 0, 0cm, 0cm}, clip]{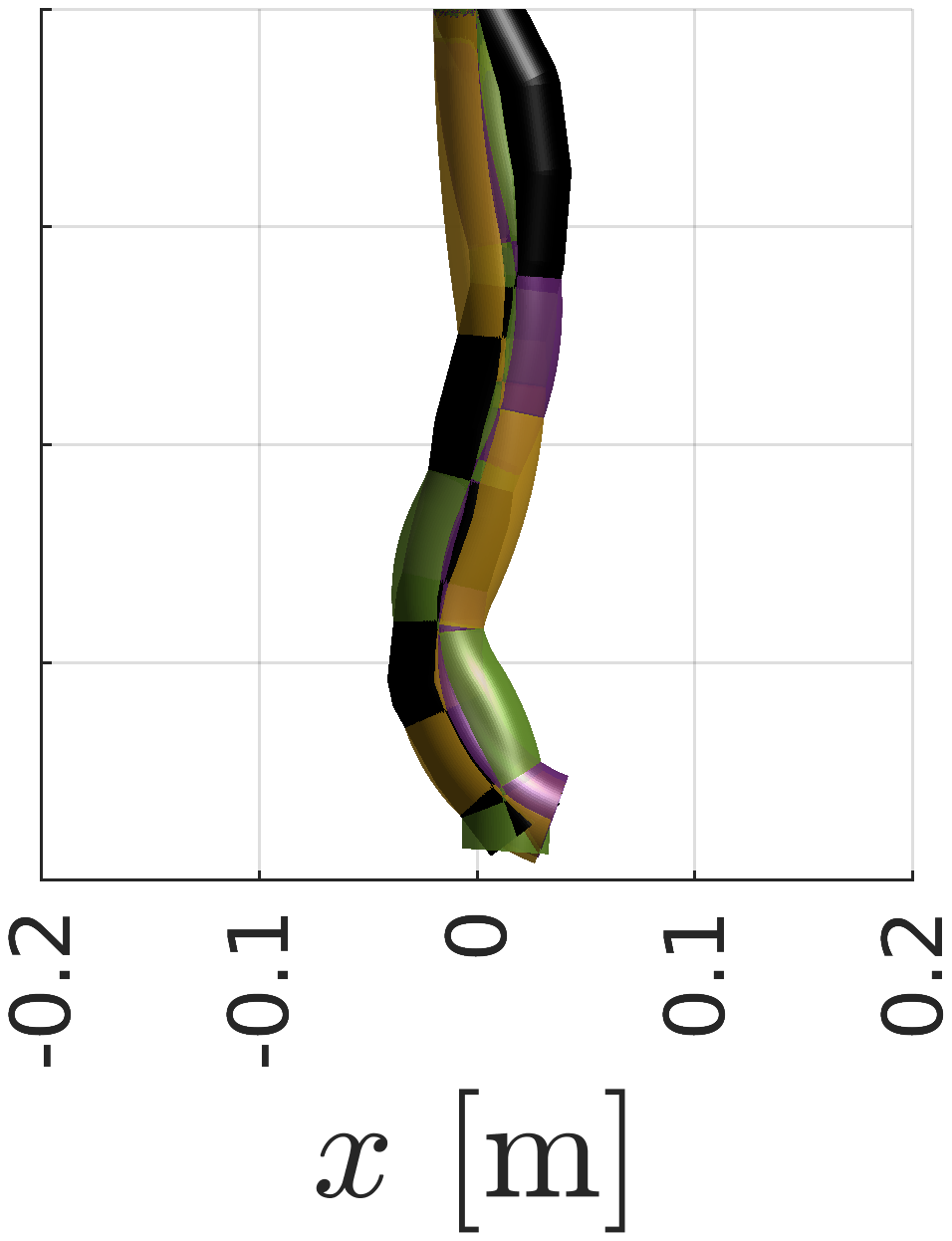}
        \label{fig:mode 3:time 4}
    }\hspace{-1cm}\hfill
    \subfigure[{$t = 0.7 T_{3}$}]{
        \includegraphics[height = 86px, trim={0cm, 0, 0cm, 0cm}, clip]{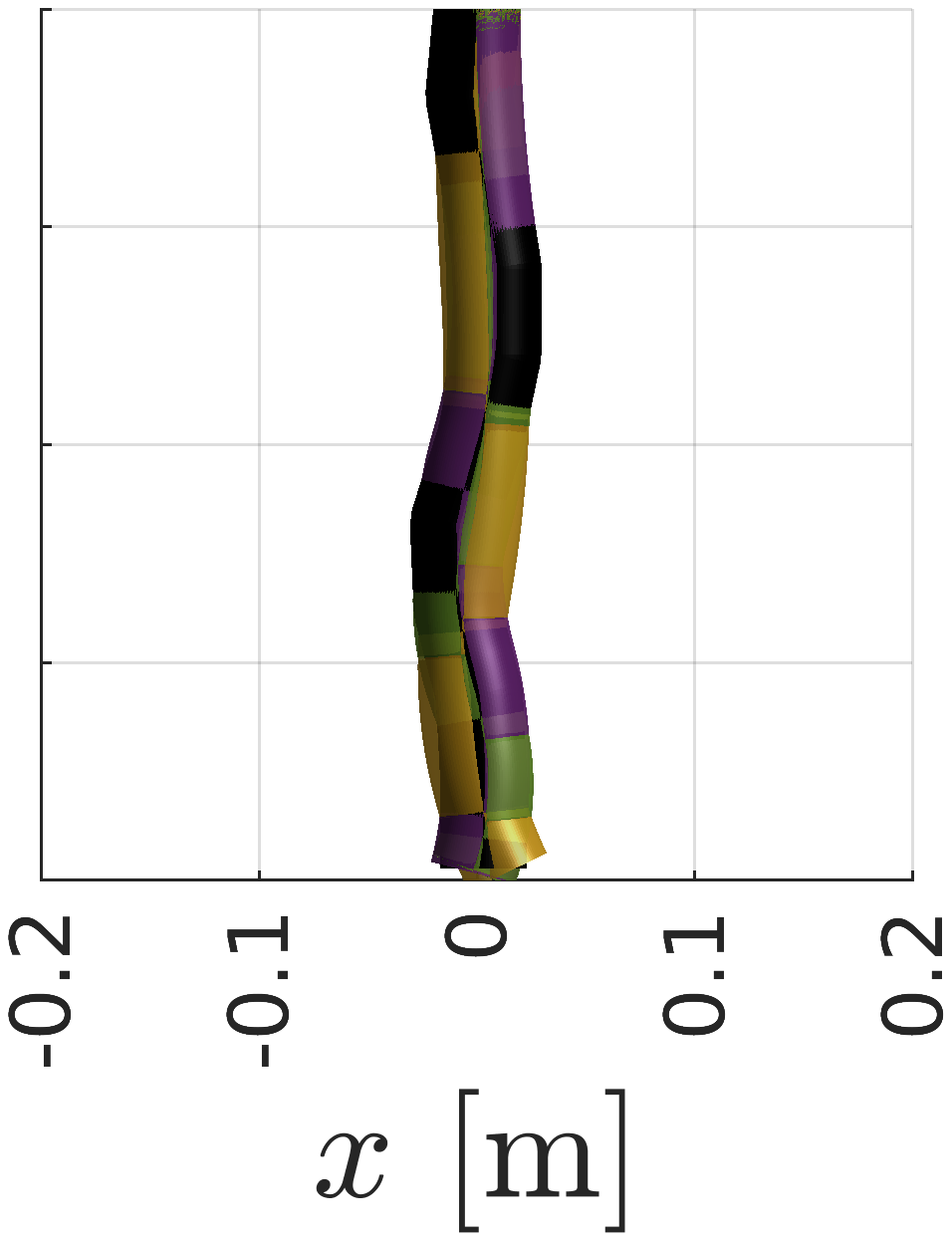}
        \label{fig:mode 3:time 5}
    }\hspace{-1cm}\hfill
    \subfigure[{$t = 0.8 T_{3}$}]{
        \includegraphics[height = 86px, trim={0cm, 0, 0cm, 0cm}, clip]{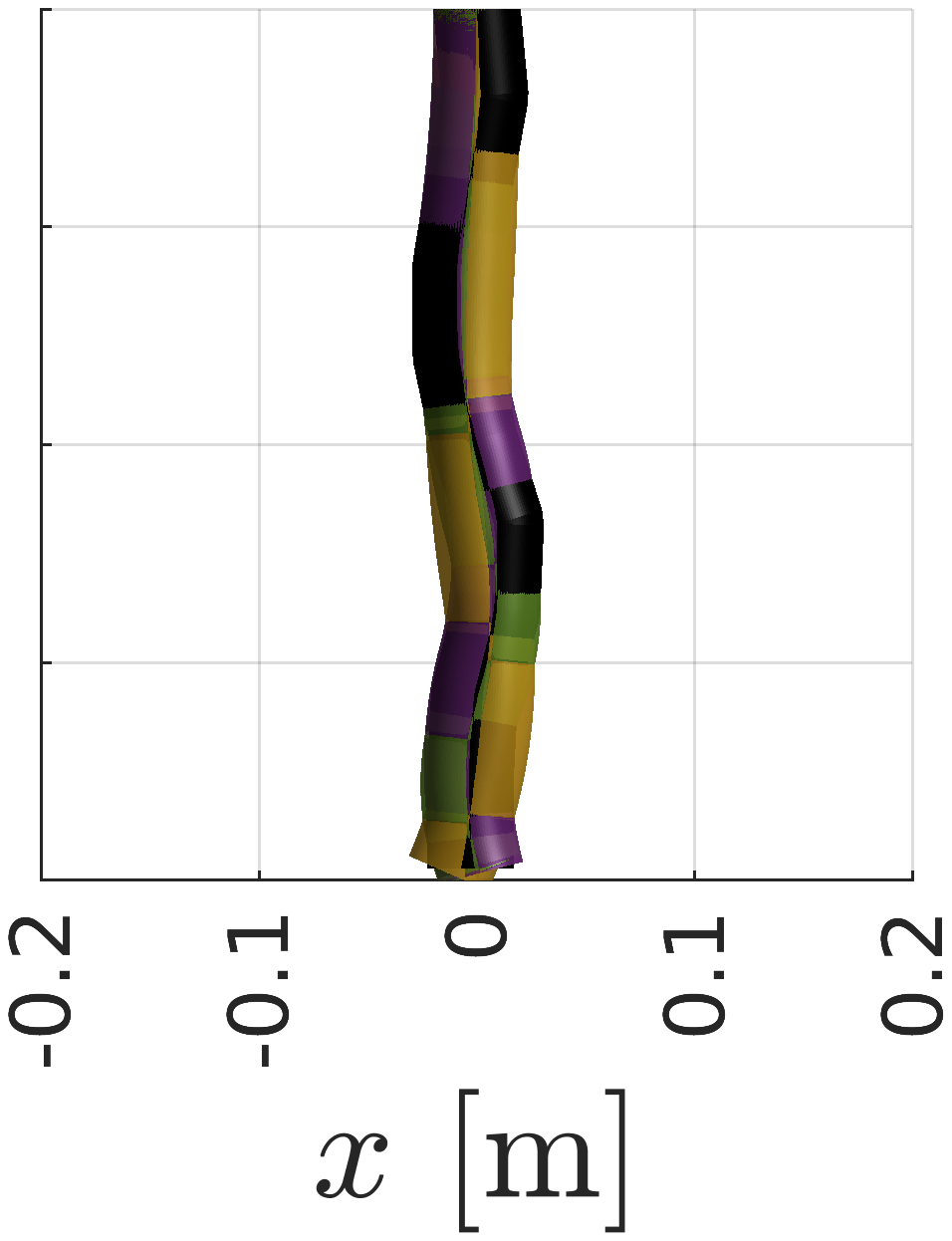}
        \label{fig:mode 3:time 6}
    }\hspace{-1cm}\hfill
    \subfigure[{$t = 0.9 T_{3}$}]{
        \includegraphics[height = 86px, trim={0cm, 0, 0cm, 0cm}, clip]{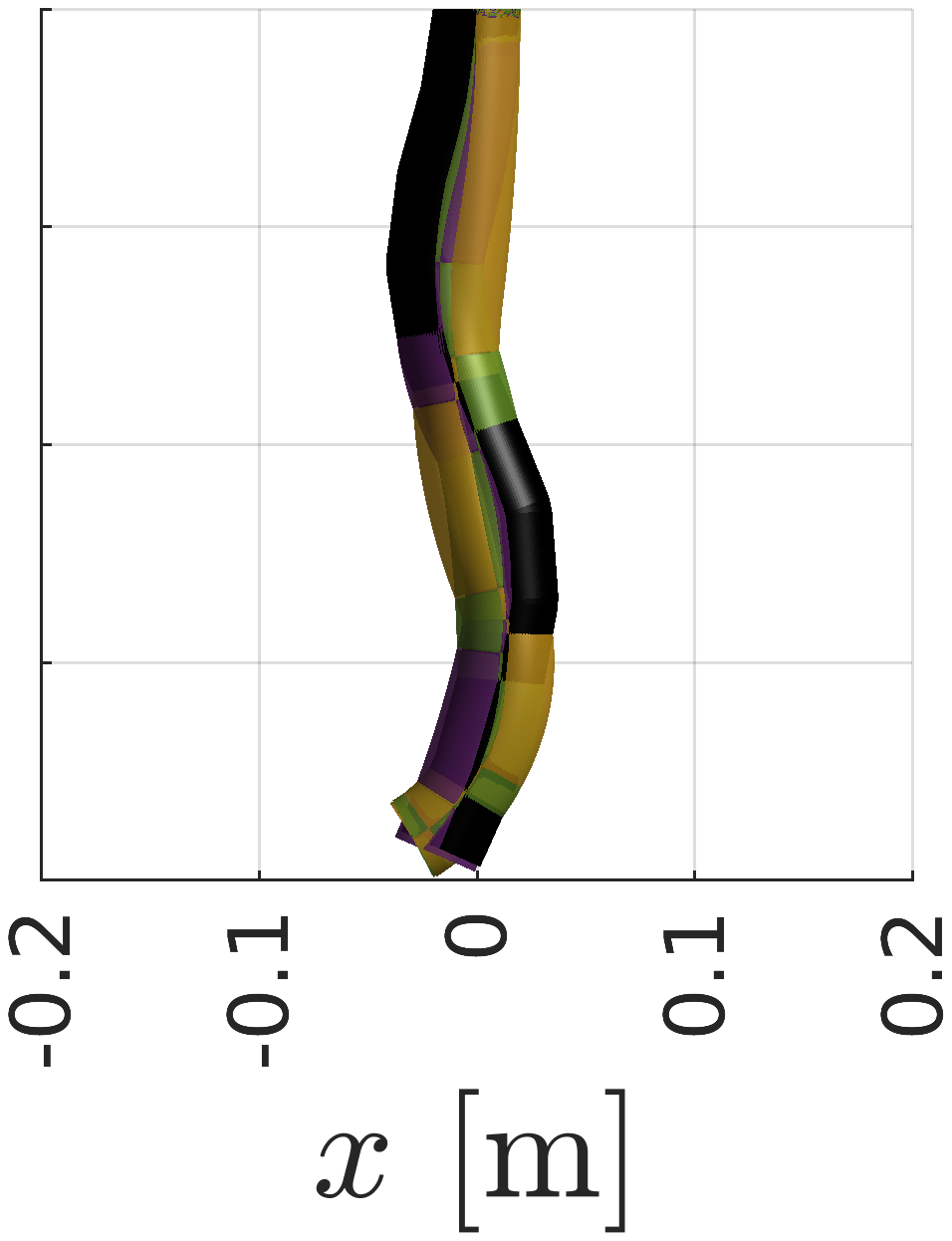}
        \label{fig:mode 3:time 7}
    }
    \caption{\small Stroboscopic plots of the robots in the workspace for the first (a)--(g), second (h)--(n) and third (o)--(u) NMs at $E_{\max} = 1~[\si{\joule}]$ and normalized time instants with respect to the mode period. Each color corresponds to a different model. 
    }
    \label{fig:strobo plot}
\end{figure*}

\begin{figure*}[!ht]
    \centering
    \subfigure[{}]{
        \includegraphics[height=14pt, trim={0 7.5cm 0 0}, clip]{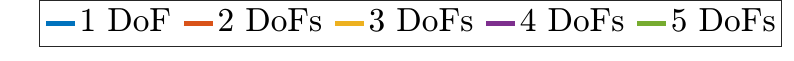}
    }\\\setcounter{subfigure}{0}\vspace{-0.8cm}
    \subfigure[{Mode $1$, $f_{x}(E, L)$}]{
        \includegraphics[width = 0.18\textwidth, height=52px, trim={0 0 0 0.9cm},clip]{matlab/ModeSolver/fig/metrics_dtw_mode_x_1}
        \label{fig:mode 1:metric 1 x}
    }\hfill
    \subfigure[{Mode $2$, $f_{x}(E, L)$}]{
        \includegraphics[width = 0.18\textwidth, height=52px, trim={0 0 0 0.9cm},clip]{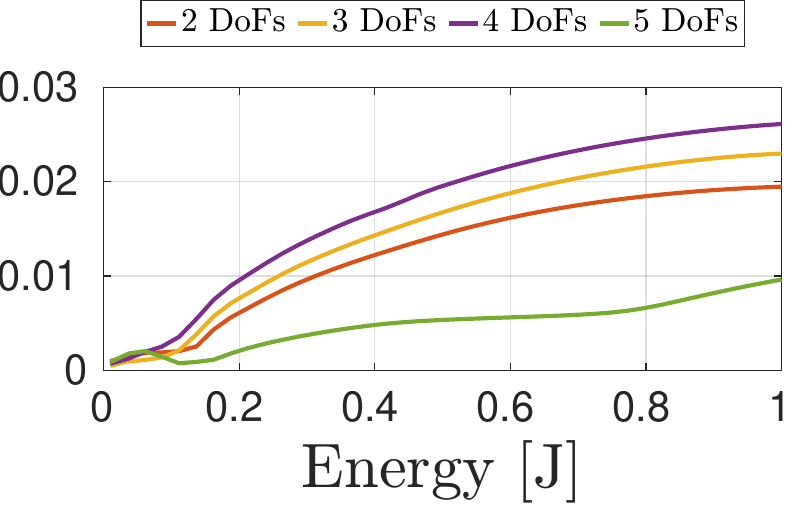}
        \label{fig:mode 2:metric 1 x}
    }\hfill
    \subfigure[{Mode $3$, $f_{x}(E, L)$}]{
        \includegraphics[width = 0.18\textwidth, height=52px, trim={0 0 0 0.9cm},clip]{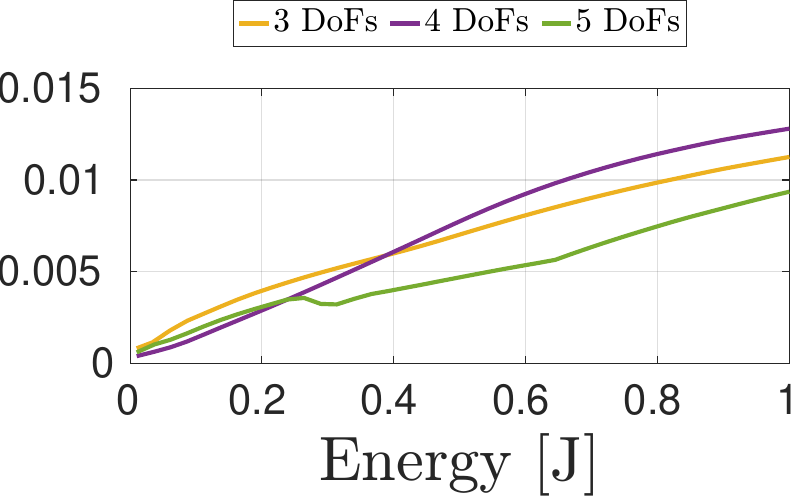}
        \label{fig:mode 3:metric 1 x}
    }\hfill
    \subfigure[{Mode $4$, $f_{x}(E, L)$}]{
        \includegraphics[width = 0.18\textwidth, height=52px, trim={0 0 0 0.9cm},clip]{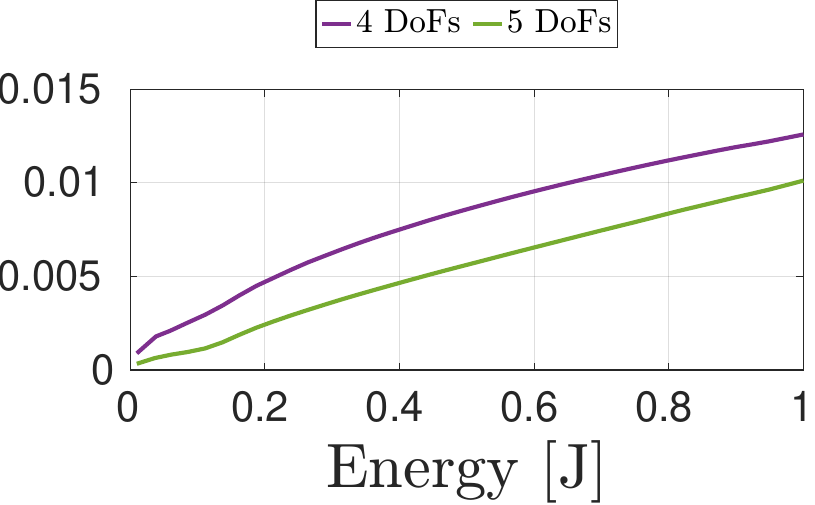}
        \label{fig:mode 4:metric 1 x}
    }\hfill
    \subfigure[{Mode $5$, $f_{x}(E, L)$}]{
        \includegraphics[width = 0.18\textwidth, height=52px, trim={0 0 0 0.9cm},clip]{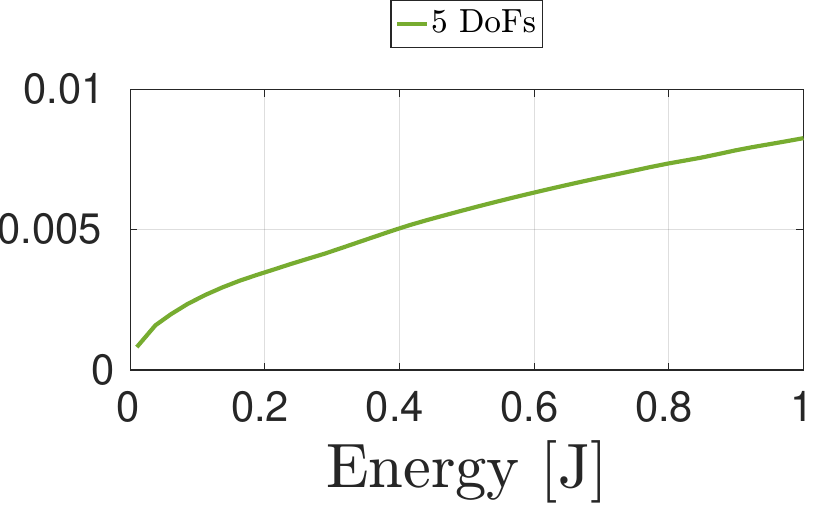}
        \label{fig:mode 5:metric 1 x}
    }\\
        \subfigure[{Mode $1$, $f_{z}(E, L)$}]{
        \includegraphics[width = 0.18\textwidth, height=52px, trim={0 0 0 0.9cm},clip]{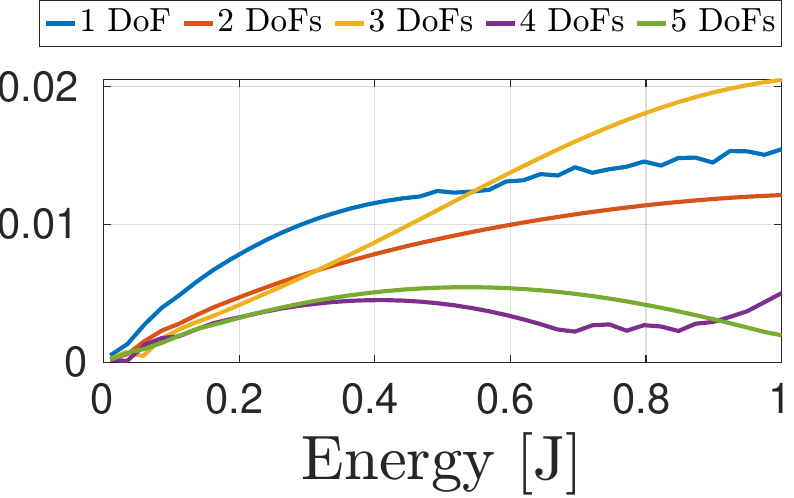}
        \label{fig:mode 1:metric 1 z}
    }\hfill
    \subfigure[{Mode $2$, $f_{z}(E, L)$}]{
        \includegraphics[width = 0.18\textwidth, height=52px, trim={0 0 0 0.9cm},clip]{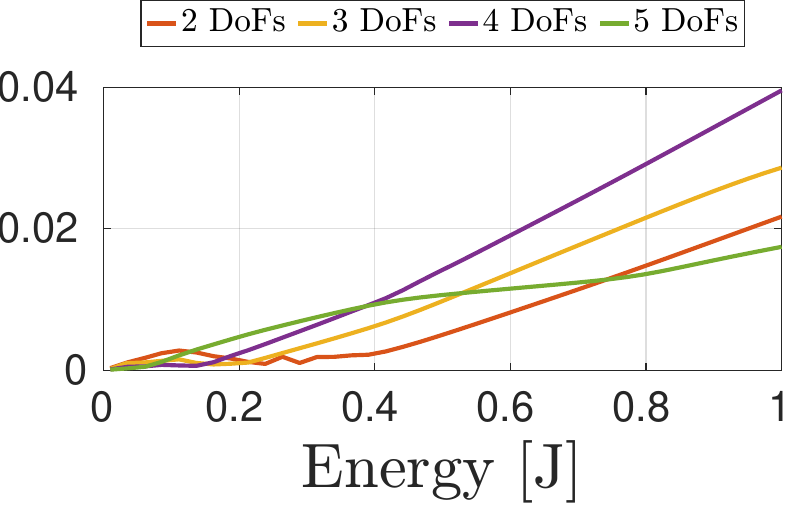}
        \label{fig:mode 2:metric 1 z}
    }\hfill
    \subfigure[{Mode $3$, $f_{z}(E, L)$}]{
        \includegraphics[width = 0.18\textwidth, height = 52px, trim={0 0 0 0.9cm},clip]{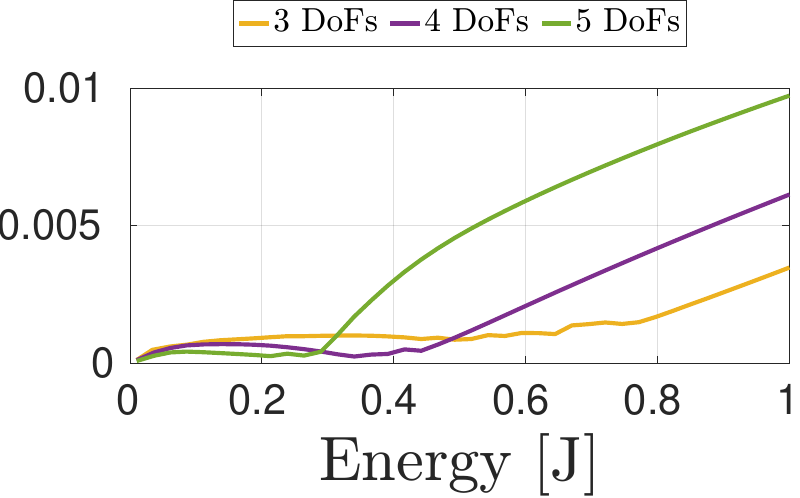}
        \label{fig:mode 3:metric 1 z}
    }\hfill
    \subfigure[{Mode $4$, $f_{z}(E, L)$}]{
        \includegraphics[width = 0.18\textwidth, height=55px, trim={0 0 0 0.9cm},clip]{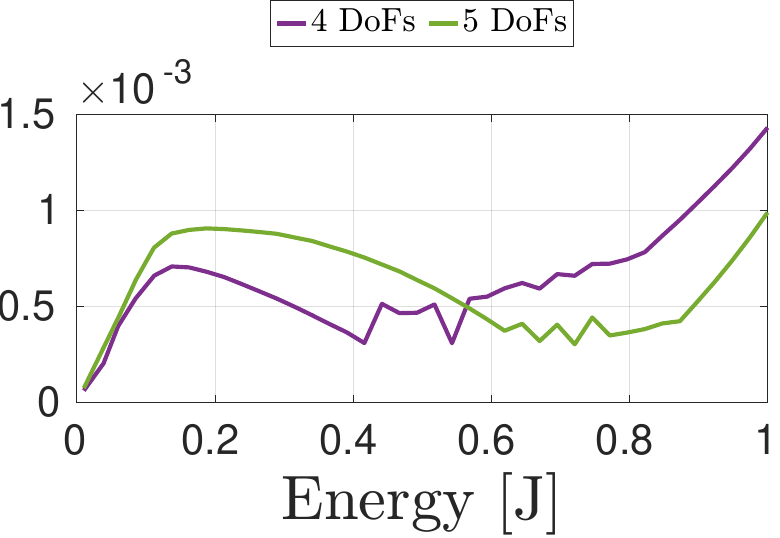}
        \label{fig:mode 4:metric 1 z}
    }\hfill
    \subfigure[{Mode $5$, $f_{z}(E, L)$}]{
        \includegraphics[width = 0.18\textwidth, height=55px, trim={0 0 0 0.9cm},clip]{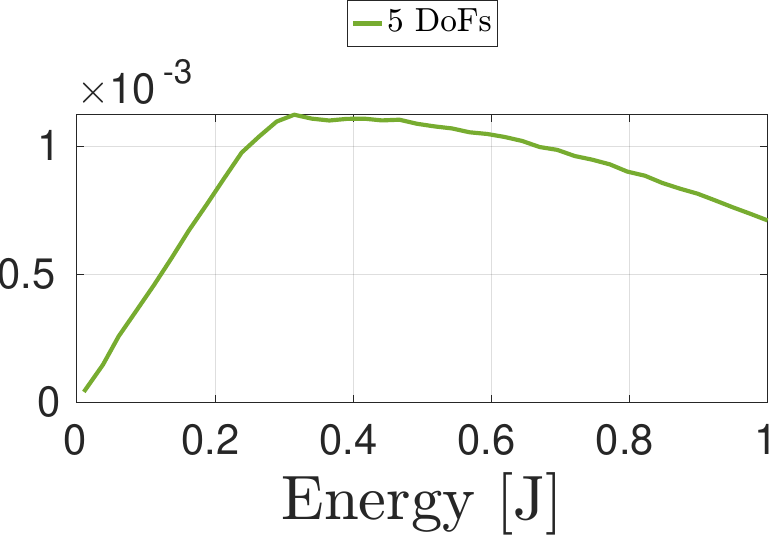}
        \label{fig:mode 5:metric 1 z}
    }
    \caption{\small Energy evolution of the modal Fréchet distance $\fv(E, L)$ for the $x$ (a)--(e) and $z$ (f)--(j) tip position.
    }
    \label{fig:f measure}
\end{figure*}

\begin{figure*}[!ht]
    \centering
    \subfigure[{}]{
        \includegraphics[height=14pt, trim={0 7.5cm 0 0}, clip]{matlab/ModeSolver/fig/metrics_dtw_mode_x_1}
    }\\\setcounter{subfigure}{0}\vspace{-0.89cm}
    \subfigure[{Mode $1$, $F_{x}(E)$}]{
        \includegraphics[width = 0.18\textwidth, height=52px, trim={0 0 0 0.9cm},clip]{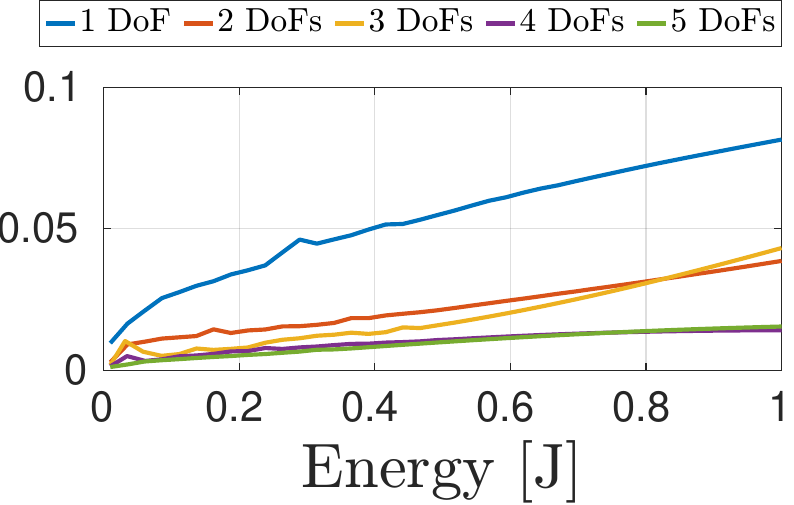}
        \label{fig:mode 1:int metric 1 x}
    }\hfill
    \subfigure[{Mode $2$, $F_{x}(E)$}]{
        \includegraphics[width = 0.18\textwidth, height=52px, trim={0 0 0 0.9cm},clip]{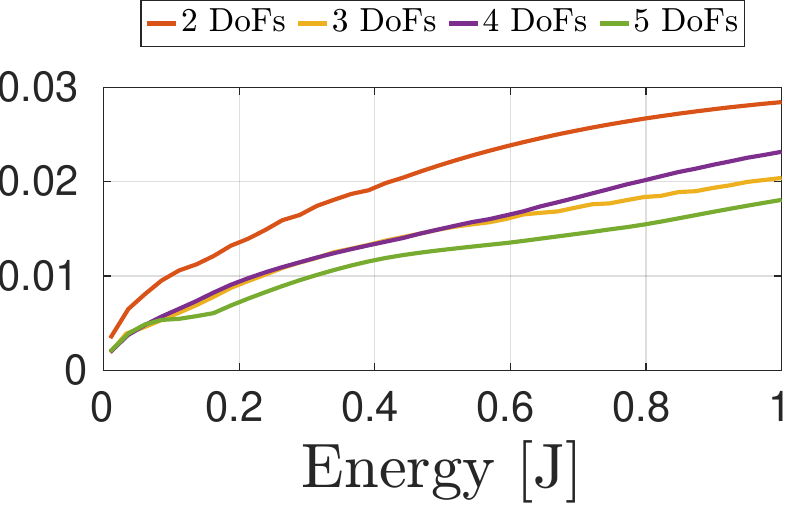}
        \label{fig:mode 2:int metric 1 x}
    }\hfill
    \subfigure[{Mode $3$, $F_{x}(E)$}]{
        \includegraphics[width = 0.18\textwidth, height=52px, trim={0 0 0 0.9cm},clip]{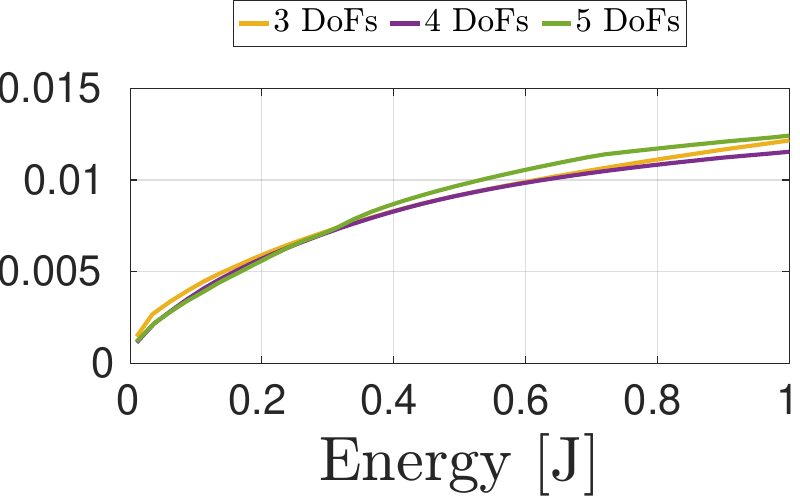}
        \label{fig:mode 3:int metric 1 x}
    }\hfill
    \subfigure[{Mode $4$, $F_{x}(E)$}]{
        \includegraphics[width = 0.18\textwidth, height=50px, trim={0 0 0 0.9cm},clip]{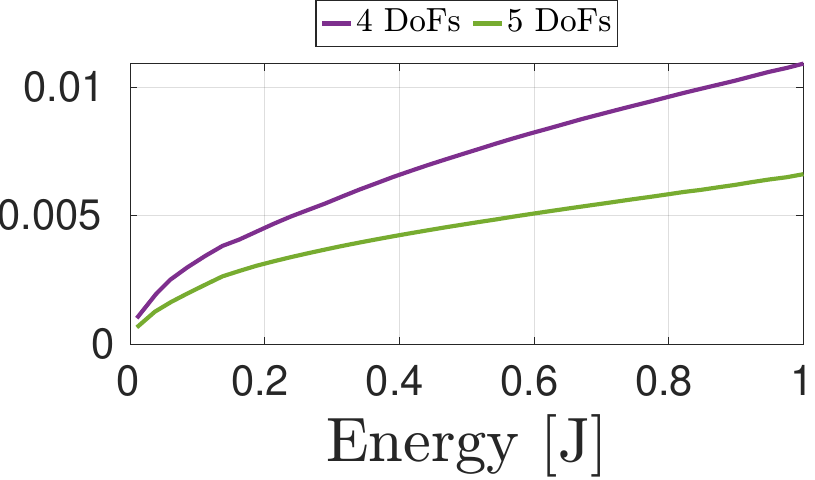}
        \label{fig:mode 4:int metric 1 x}
    }\hfill
    \subfigure[{Mode $5$, $F_{x}(E)$}]{
        \includegraphics[width = 0.18\textwidth, height = 56px, trim={0 0 0 0.9cm},clip]{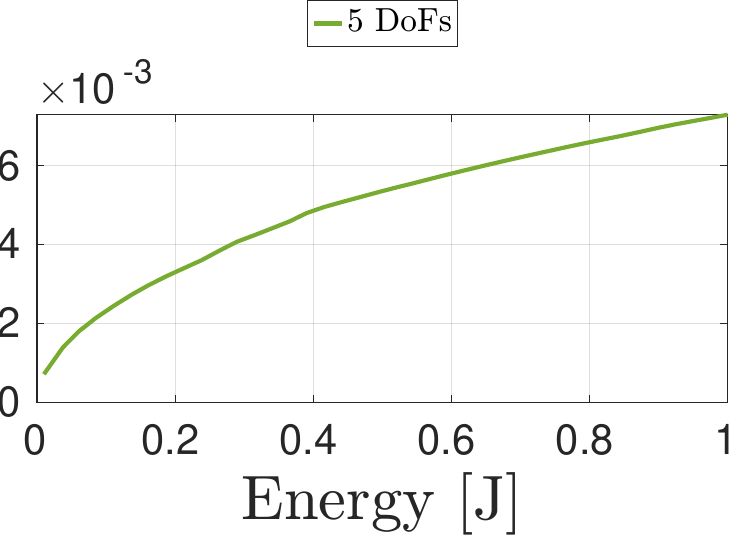}
        \label{fig:mode 5:int metric 1 x}
    }\\
        \subfigure[{Mode $1$, $F_{z}(E)$}]{
        \includegraphics[width = 0.18\textwidth, height = 50px, trim={0 0 0 0.9cm},clip]{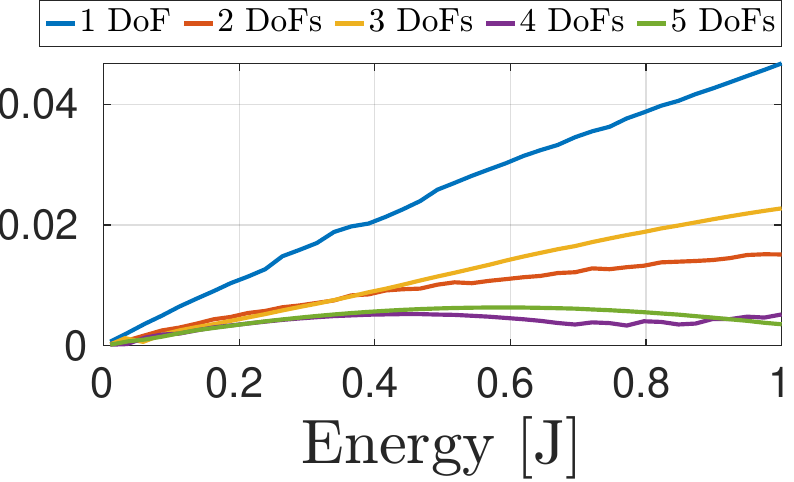}
        \label{fig:mode 1:int metric 1 z}
    }\hfill
    \subfigure[{Mode $2$, $F_{z}(E)$}]{
        \includegraphics[width = 0.18\textwidth, height = 52px, trim={0 0 0 0.9cm},clip]{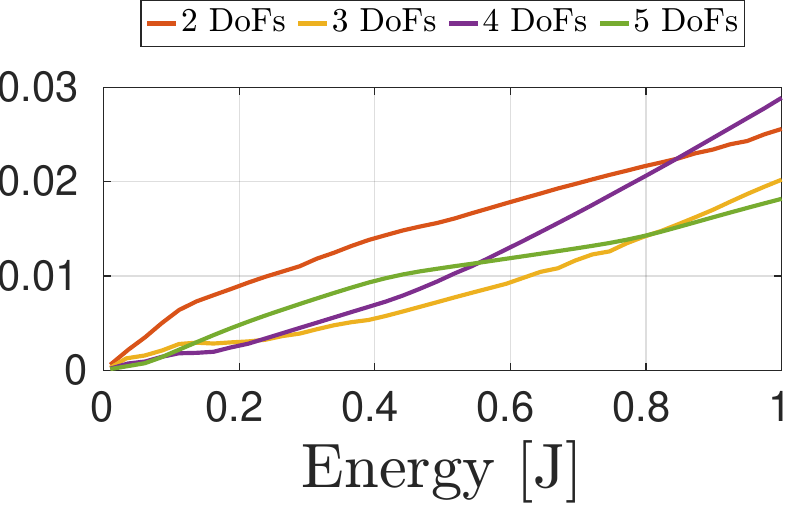}
        \label{fig:mode 2:int metric 1 z}
    }\hfill
    \subfigure[{Mode $3$, $F_{z}(E)$}]{
        \includegraphics[width = 0.18\textwidth, height = 52px, trim={0 0 0 0.9cm},clip]{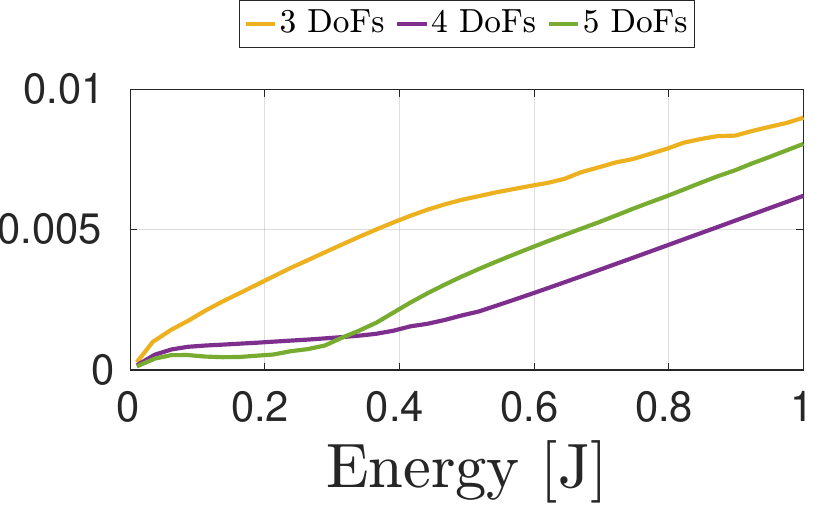}
        \label{fig:mode 3:int metric 1 z}
    }\hfill
    \subfigure[{Mode $4$, $F_{z}(E)$}]{
        \includegraphics[width = 0.18\textwidth, height = 52px, trim={0 0 0 0.9cm},clip]{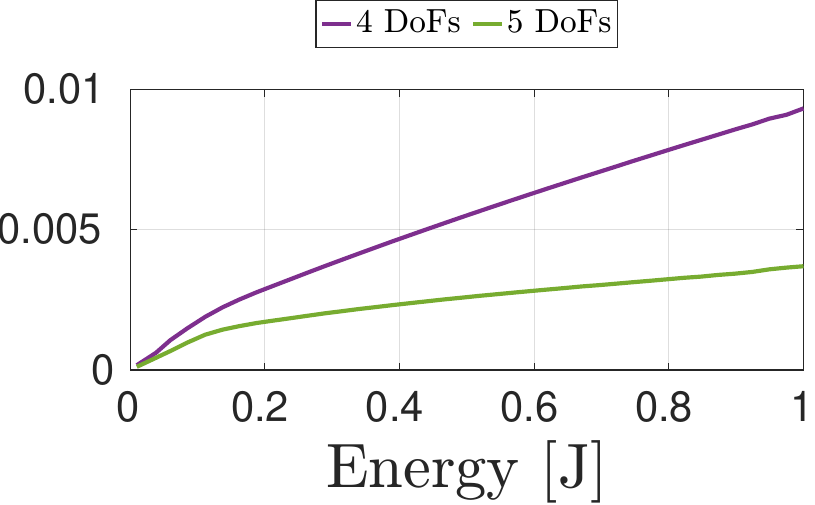}
        \label{fig:mode 4:int metric 1 z}
    }\hfill
    \subfigure[{Mode $5$, $F_{z}(E)$}]{
        \includegraphics[width = 0.18\textwidth, height = 55px, trim={0 0 0 0.9cm},clip]{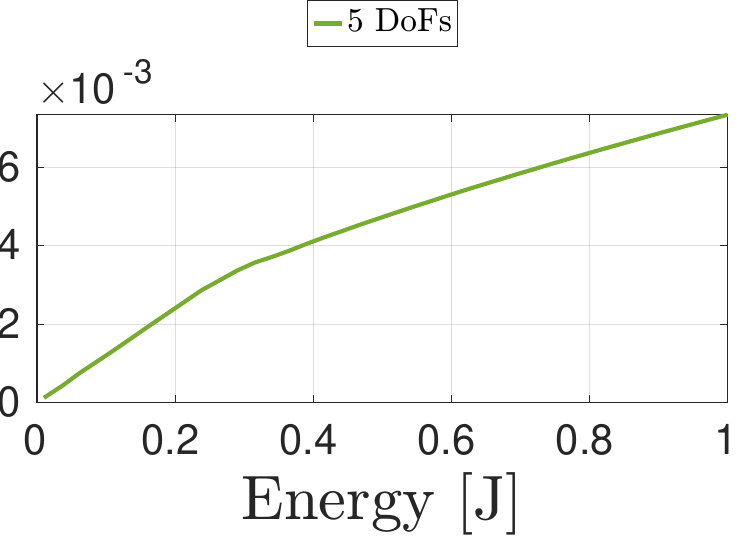}
        \label{fig:mode 5:int metric 1 z}
    }
    \caption{\small Energy evolution of the modal integral Fréchet distance $\Fv(E)$ for the $x$ (a)--(e) and $z$ (f)--(j) backbone position. 
    }
    \label{fig:f int measure}
\end{figure*}

\begin{figure*}[!ht]
    \centering
    \subfigure[{}]{
        \includegraphics[height=14pt, trim={0 7.5cm 0 0}, clip]{matlab/ModeSolver/fig/metrics_dtw_mode_x_1}
    }\\\setcounter{subfigure}{0}\vspace{-0.89cm}
    \subfigure[{Mode $1$, $g_{x}(E, L)$}]{
        \includegraphics[width = 0.18\textwidth, height=54px, trim={0 0 0 0.9cm},clip]{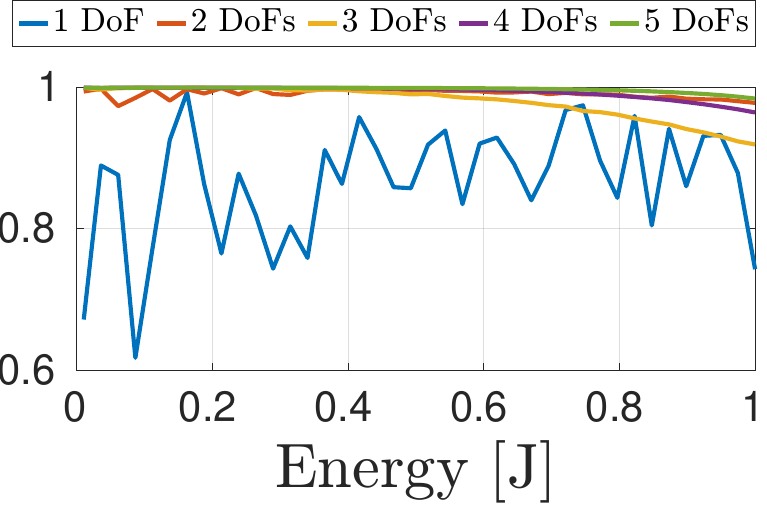}
        \label{fig:mode 1:metric 2 x}
    }\hfill
    \subfigure[{Mode $2$, $g_{x}(E, L)$}]{
        \includegraphics[width = 0.18\textwidth, height=54px, trim={0 0 0 0.9cm},clip]{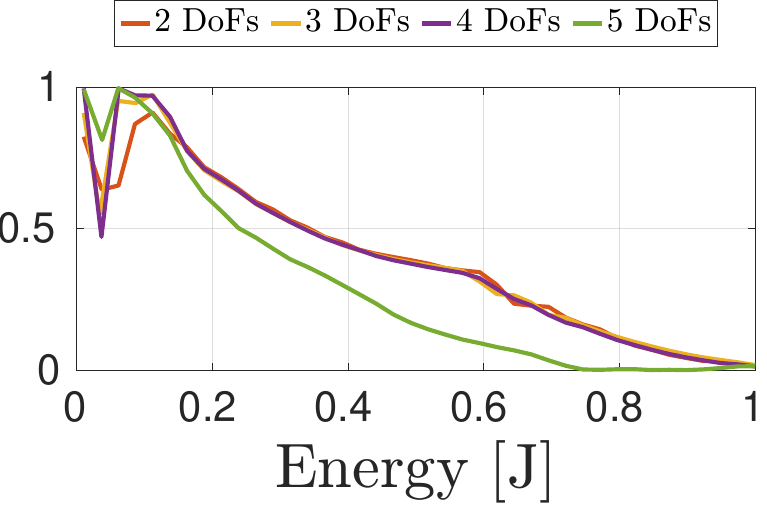}
        \label{fig:mode 2:metric 2 x}
    }\hfill
    \subfigure[{Mode $3$, $g_{x}(E, L)$}]{
        \includegraphics[width = 0.18\textwidth, height=54px, trim={0 0 0 0.9cm},clip]{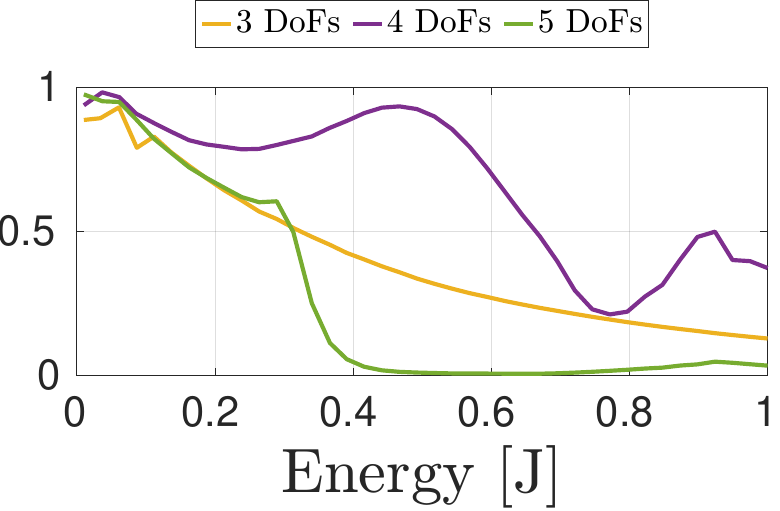}
        \label{fig:mode 3:metric 2 x}
    }\hfill
    \subfigure[{Mode $4$, $g_{x}(E, L)$}]{
        \includegraphics[width = 0.18\textwidth, height=54px, trim={0 0 0 0.9cm},clip]{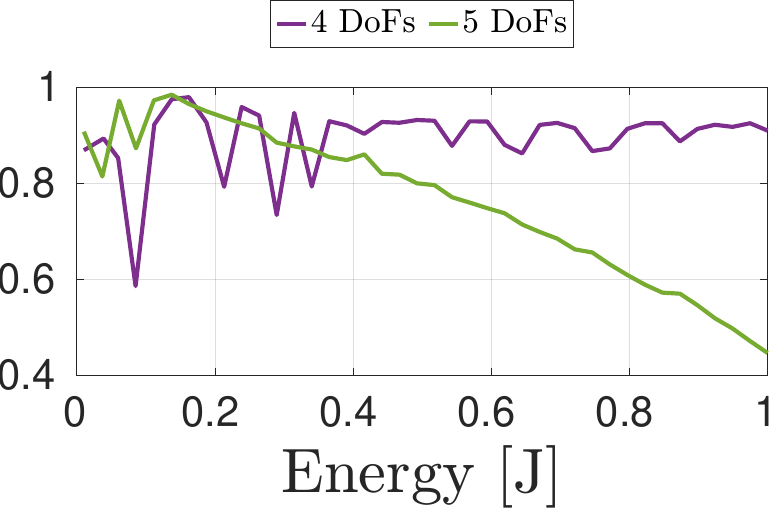}
        \label{fig:mode 4:metric 2 x}
    }\hfill
    \subfigure[{Mode $5$, $g_{x}(E, L)$}]{
        \includegraphics[width = 0.18\textwidth, height=54px, trim={0 0 0 0.9cm},clip]{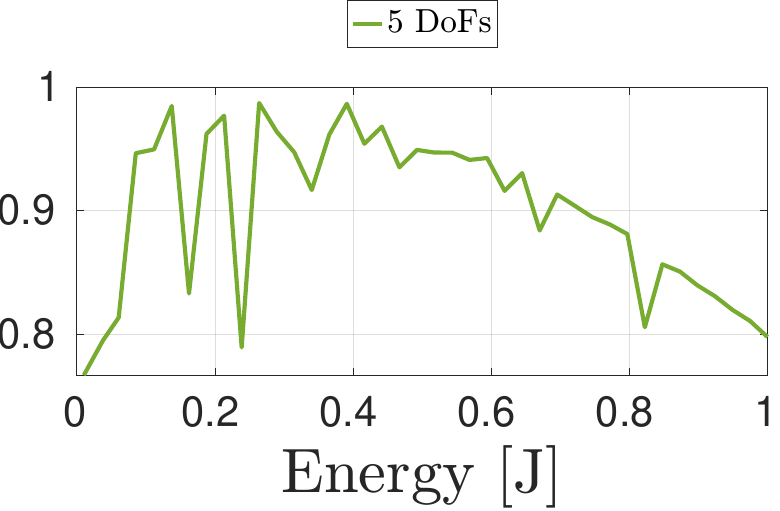}
        \label{fig:mode 5:metric 2 x}
    }\\
        \subfigure[{Mode $1$, $g_{x}(E, L)$}]{
        \includegraphics[width = 0.18\textwidth, height=54px, trim={0 0 0 0.9cm},clip]{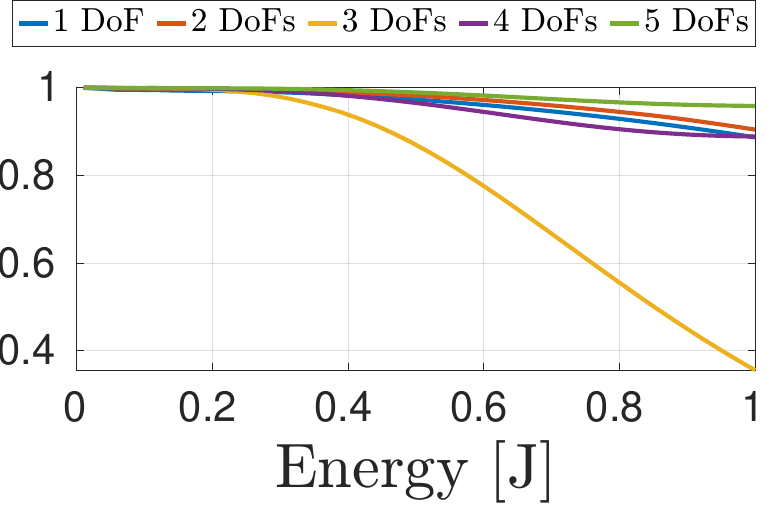}
        \label{fig:mode 1:metric 2 z}
    }\hfill
    \subfigure[{Mode $2$, $g_{z}(E, L)$}]{
        \includegraphics[width = 0.18\textwidth, height=54px, trim={0 0 0 0.9cm},clip]{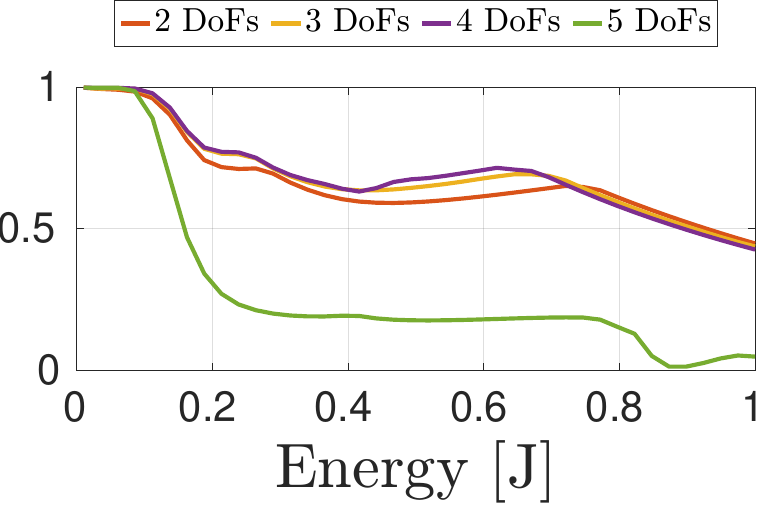}
        \label{fig:mode 2:metric 2 z}
    }\hfill
    \subfigure[{Mode $3$, $g_{z}(E, L)$}]{
        \includegraphics[width = 0.18\textwidth, height=54px, trim={0 0 0 0.9cm},clip]{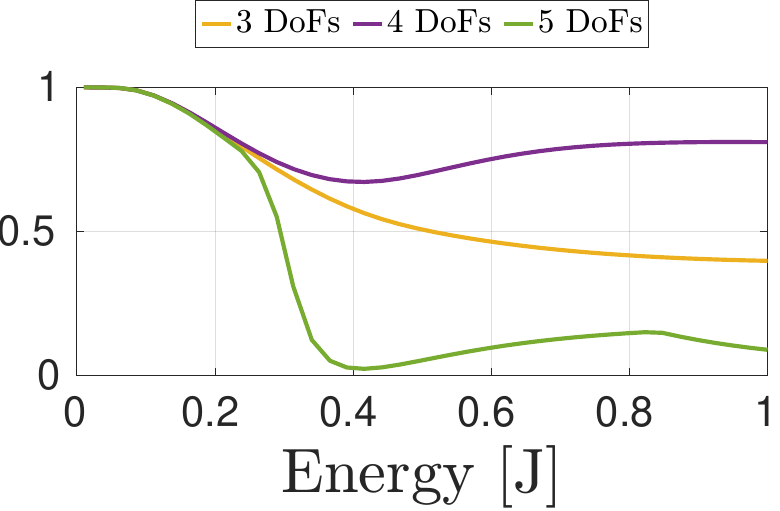}
        \label{fig:mode 3:metric 2 z}
    }\hfill
    \subfigure[{Mode $4$, $g_{z}(E, L)$}]{
        \includegraphics[width = 0.18\textwidth, height=54px, trim={0 0 0 0.9cm},clip]{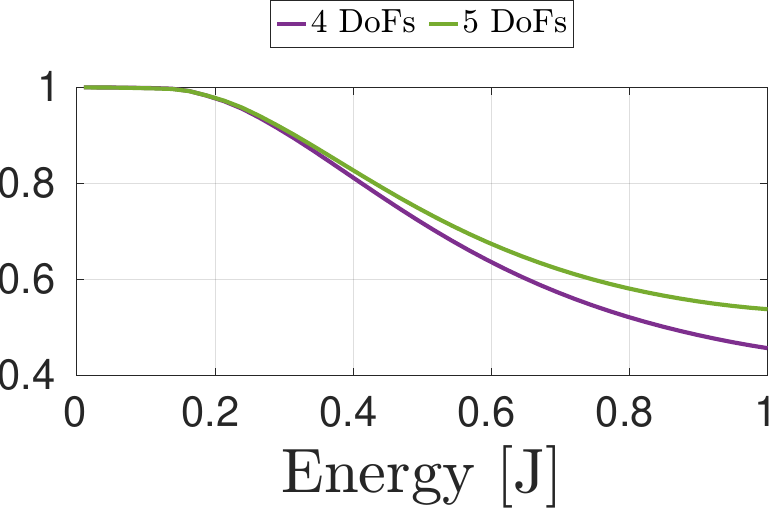}
        \label{fig:mode 4:metric 2 z}
    }\hfill
    \subfigure[{Mode $5$, $g_{z}(E, L)$}]{
        \includegraphics[width = 0.18\textwidth, height=54px, trim={0 0 0 0.9cm},clip]{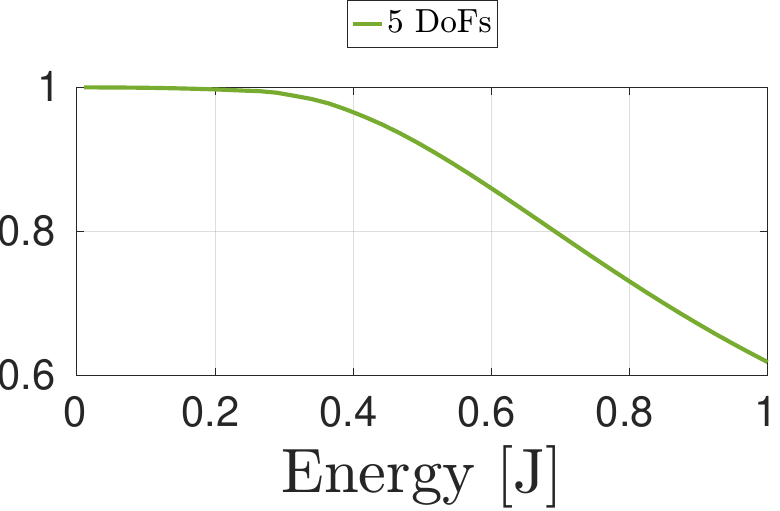}
        \label{fig:mode 5:metric 2 z}
    }
    \caption{\small Energy evolution of the modal coherence $\gv(E, L)$ for the $x$ (a)--(e) and $z$ (f)--(j) tip position.  
    }
    \label{fig:g measure}
\end{figure*}

\begin{figure*}[!ht]
    \centering
    \subfigure[{}]{
        \includegraphics[height=14pt, trim={0 7.5cm 0 0}, clip]{matlab/ModeSolver/fig/metrics_dtw_mode_x_1}
    }\\\setcounter{subfigure}{0}\vspace{-0.89cm}
    \subfigure[{Mode $1$, $G_{x}(E)$}]{
        \includegraphics[width = 0.18\textwidth, height=54px, trim={0 0 0 0.9cm},clip]{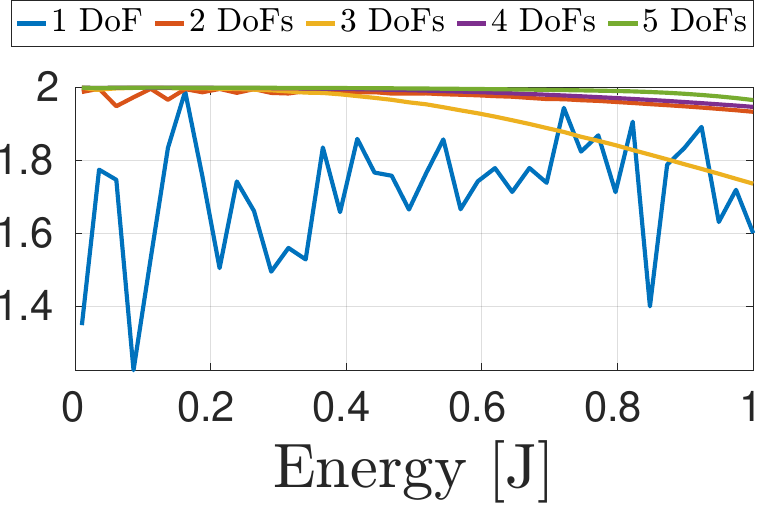}
        \label{fig:mode 1:int metric 2 x}
    }\hfill
    \subfigure[{Mode $2$, $G_{x}(E)$}]{
        \includegraphics[width = 0.18\textwidth, height=54px, trim={0 0 0 0.9cm},clip]{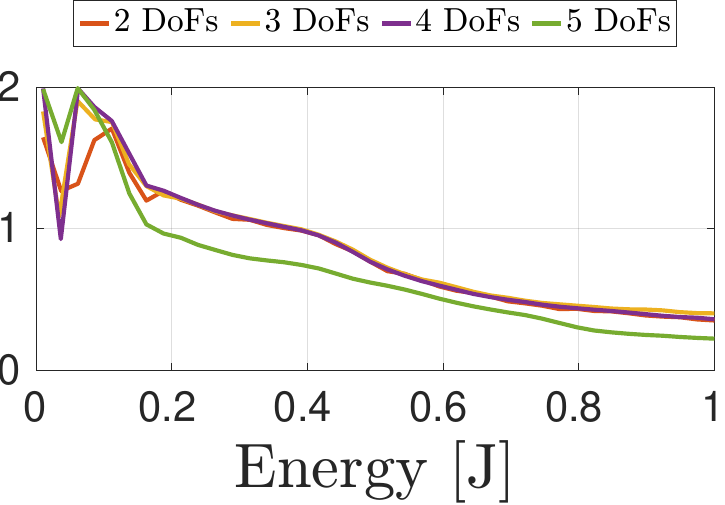}
        \label{fig:mode 2:int metric 2 x}
    }\hfill
    \subfigure[{Mode $3$, $G_{x}(E)$}]{
        \includegraphics[width = 0.18\textwidth, height=54px, trim={0 0 0 0.9cm},clip]{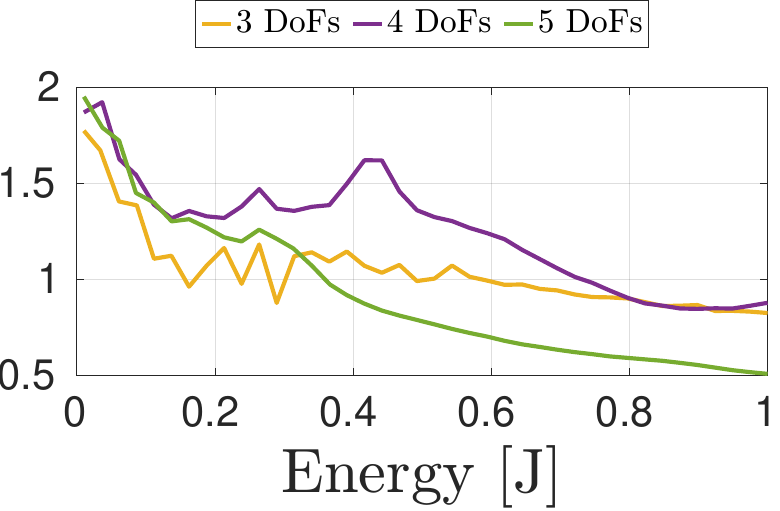}
        \label{fig:mode 3:int metric 2 x}
    }\hfill
    \subfigure[{Mode $4$, $G_{x}(E)$}]{
        \includegraphics[width = 0.18\textwidth, height=54px, trim={0 0 0 0.9cm},clip]{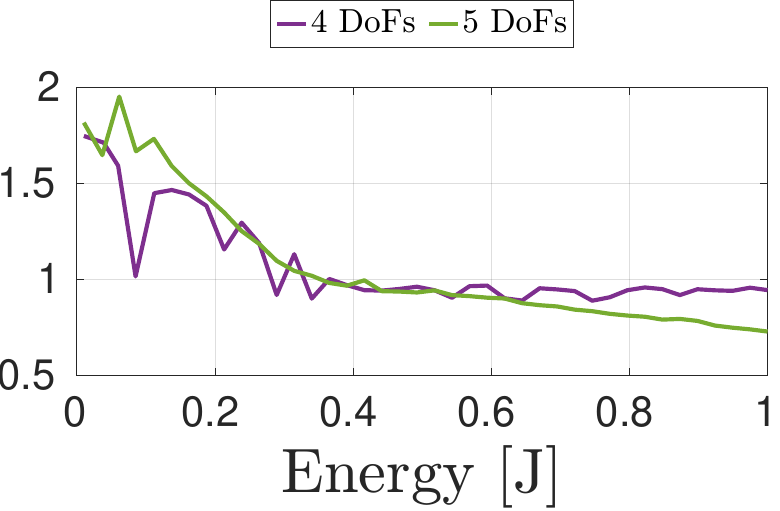}
        \label{fig:mode 4:int metric 2 x}
    }\hfill
    \subfigure[{Mode $5$, $G_{x}(E)$}]{
        \includegraphics[width = 0.18\textwidth, height=51px, trim={0 0 0 0.9cm},clip]{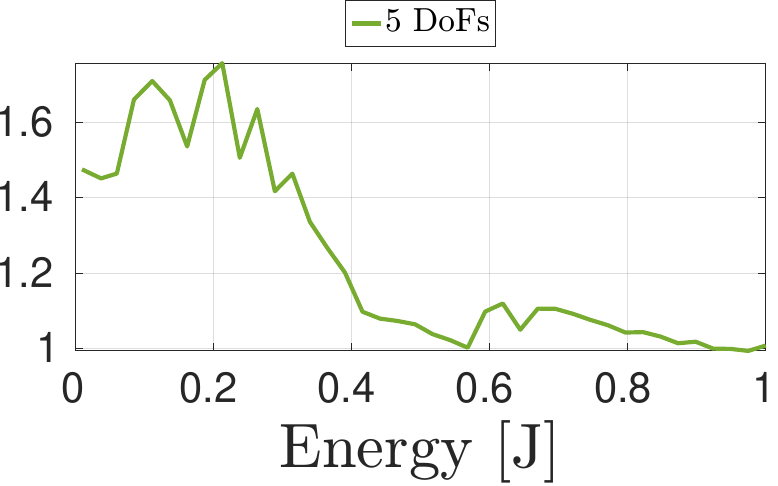}
        \label{fig:mode 5:int metric 2 x}
    }\\
        \subfigure[{Mode $1$, $G_{x}(E)$}]{
        \includegraphics[width = 0.18\textwidth, height=54px, trim={0 0 0 0.9cm},clip]{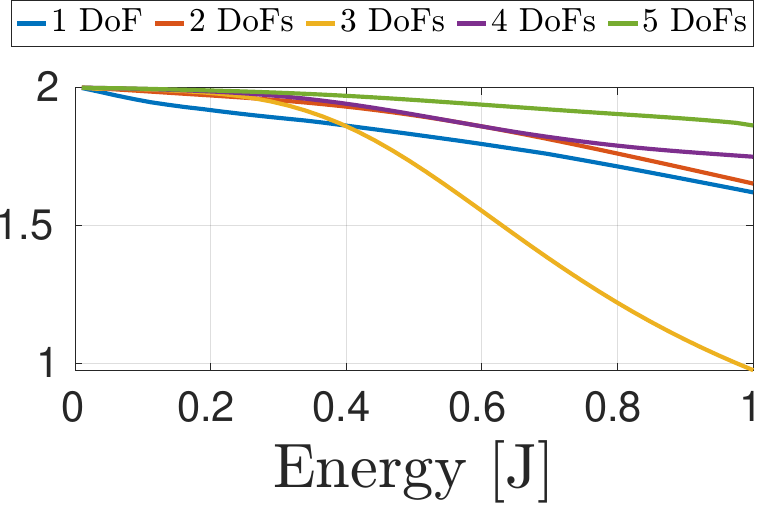}
        \label{fig:mode 1:int metric 2 z}
    }\hfill
    \subfigure[{Mode $2$, $G_{z}(E)$}]{
        \includegraphics[width = 0.18\textwidth, height=54px, trim={0 0 0 0.9cm},clip]{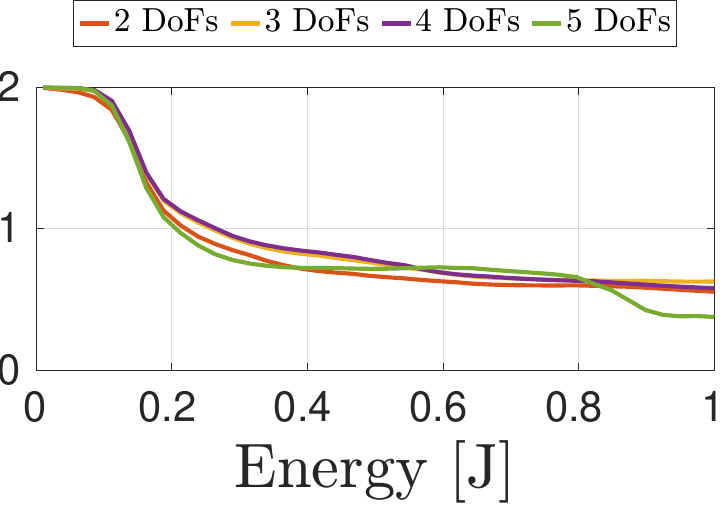}
        \label{fig:mode 2:int metric 2 z}
    }\hfill
    \subfigure[{Mode $3$, $G_{z}(E)$}]{
        \includegraphics[width = 0.18\textwidth, height=54px, trim={0 0 0 0.9cm},clip]{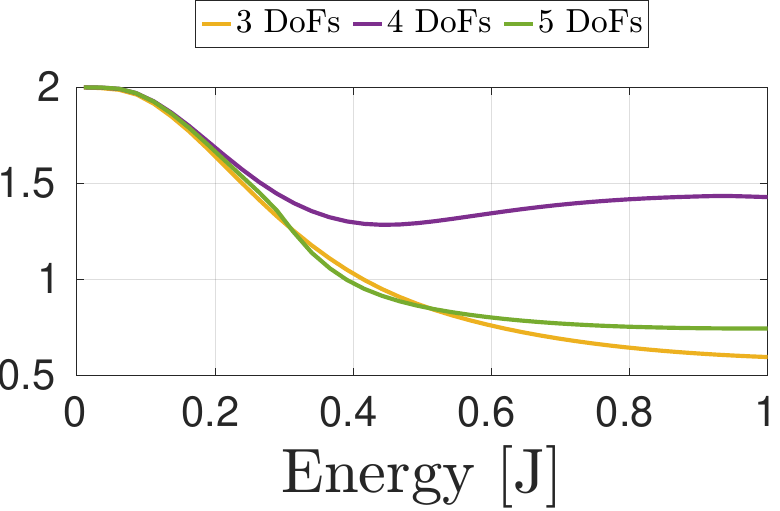}
        \label{fig:mode 3:int metric 2 z}
    }\hfill
    \subfigure[{Mode $4$, $G_{z}(E)$}]{
        \includegraphics[width = 0.18\textwidth, height=54px, trim={0 0 0 0.9cm},clip]{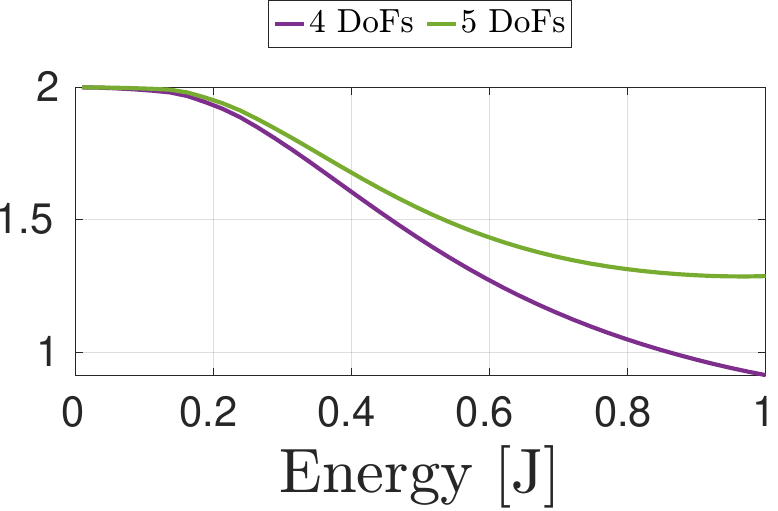}
        \label{fig:mode 4:int metric 2 z}
    }\hfill
    \subfigure[{Mode $5$, $G_{z}(E)$}]{
        \includegraphics[width = 0.18\textwidth, height=54px, trim={0 0 0 0.9cm},clip]{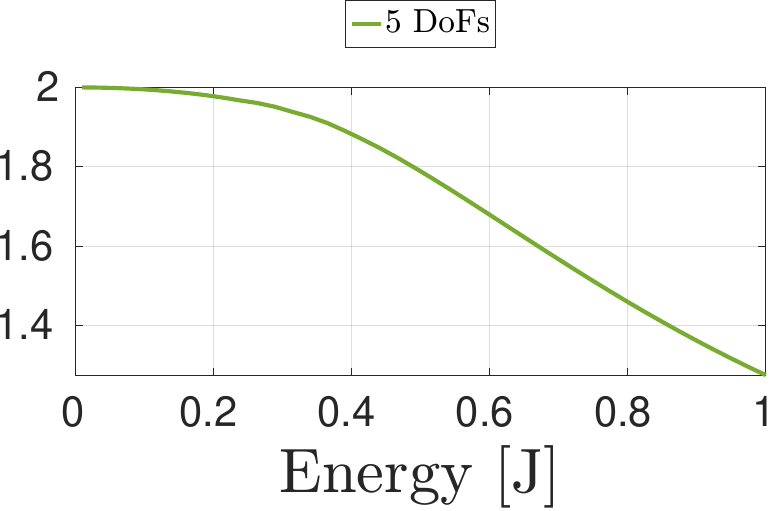}
        \label{fig:mode 5:int metric 2 z}
    }
    \caption{\small Energy evolution of the modal integral coherence $\Gm(E)$ for the $x$ (a)--(e) and $z$ (f)--(j) backbone position. 
    }
    \label{fig:g int measure}
\end{figure*}

Figure~\ref{fig:strobo plot} shows stroboscopic plots of the robots in the workspace for the first three NMs at $E_{\max}$. 
\begin{figure*}[!ht]
    \centering
    \subfigure[{}]{
        \includegraphics[height=10pt, trim={0 7cm 0 0}, clip]{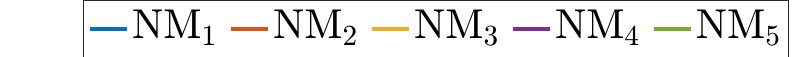}
    }\\\setcounter{subfigure}{0}\vspace{-0.77cm}
    \subfigure[{PCC $1$ DoF}]{
        \includegraphics[width = 0.31\textwidth, trim={0 0 0 1.1cm}, clip]{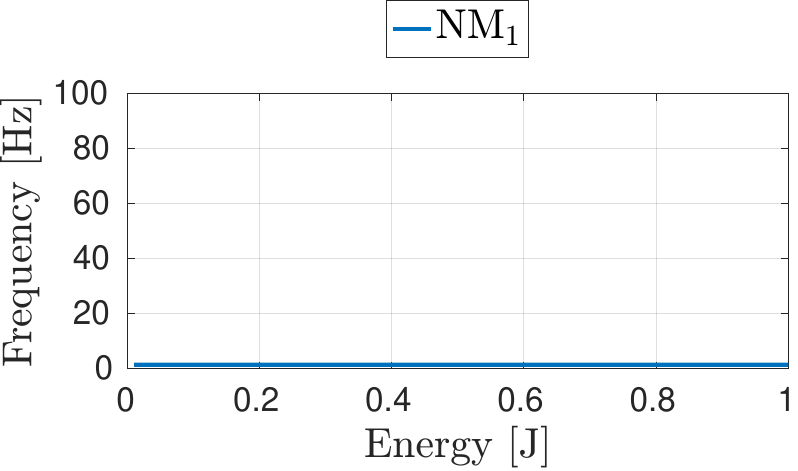}
        \label{fig:PCC 1 DOF:energy_frequency}
    }\hfill
    \subfigure[{PCC $2$ DoFs}]{
        \includegraphics[width = 0.31\textwidth, trim={0 0 0 1.1cm}, clip]{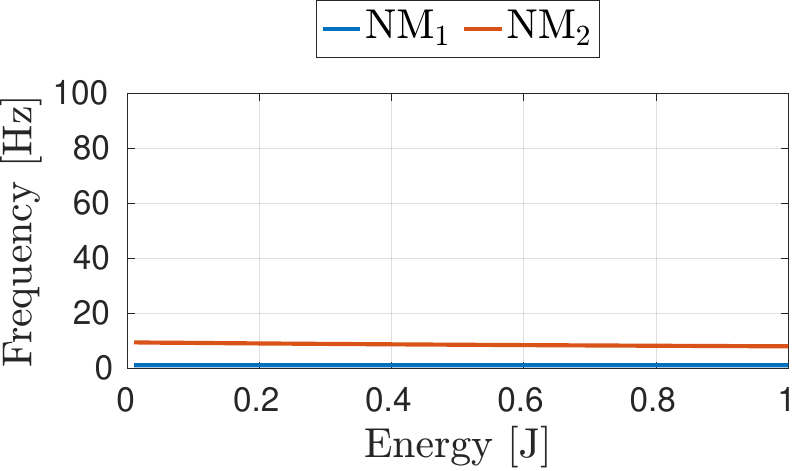}
        \label{fig:PCC 2 DOF:energy_frequency}
    }\hfill
    \subfigure[{PCC $3$ DoFs}]{
        \includegraphics[width = 0.31\textwidth, trim={0 0 0 1.1cm}, clip]{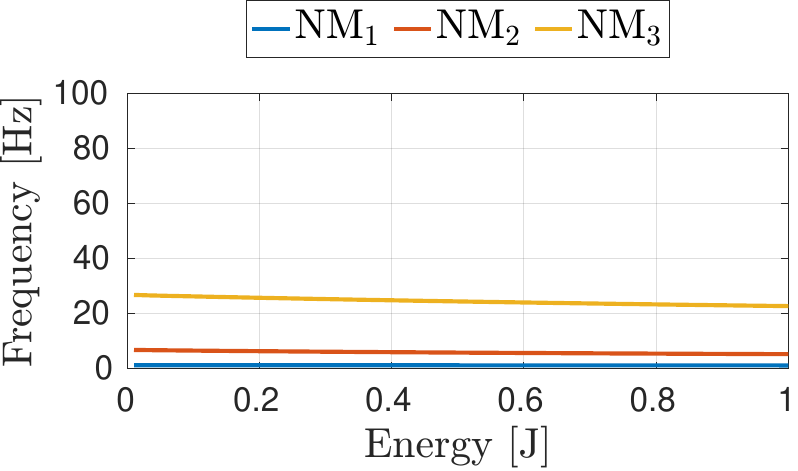}
        \label{fig:PCC 3 DOF:energy_frequency}
    }\\
    \subfigure[{PCC $4$ DoFs}]{
        \includegraphics[width = 0.31\textwidth, trim={0 0 0 1.1cm}, clip]{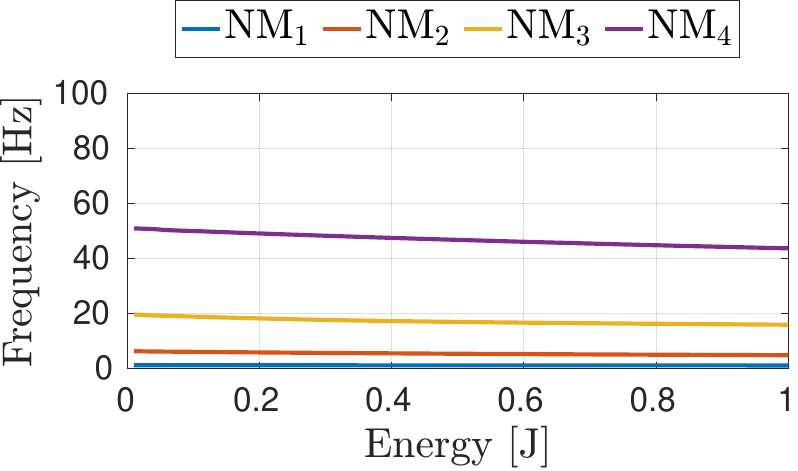}
        \label{fig:PCC 4 DOF:energy_frequency}
    }\hfill
    \subfigure[{PCC $5$ DoFs}]{
        \includegraphics[width = 0.31\textwidth, trim={0 0 0 1.1cm}, clip]{matlab/ModeSolver/fig/sim5_energy_frequency}
        \label{fig:PCC 5 DOF:energy_frequency}
    }\hfill
    \subfigure[{Rigid $10$ DoFs}]{
        \includegraphics[width = 0.31\textwidth, trim={0 0 0 1.1cm}, clip]{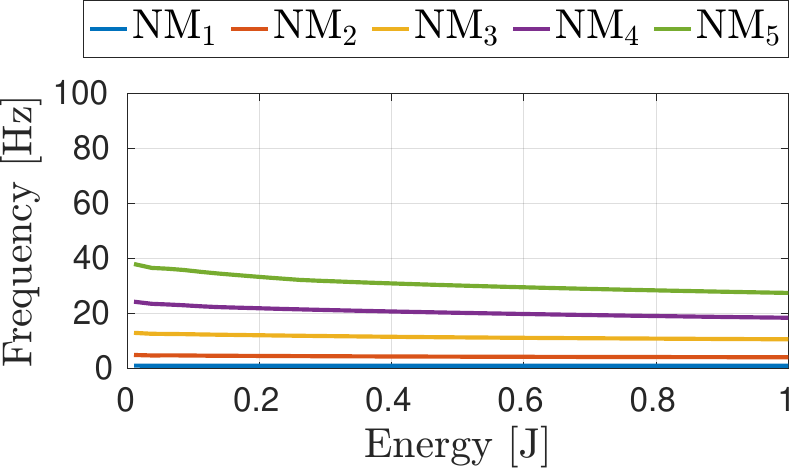}
        \label{fig:PCC 6 DOF:energy_frequency}
    }
    \caption{\small Energy-frequency relationships for the PCC~(a)--(e) and FE~(f) discretization, respectively. The models show a similar behavior only for the first two modes.
    }
    \label{fig:energy frequency}
\end{figure*}
Figure~\ref{fig:energy frequency} depicts the energy-frequency plot. Only the first two eigenmanifolds at lower frequency have trajectories with similar periods. These differences are reasonably expected to show up also in the task space, at least for the coherence. Figure~\ref{fig:f measure} reports the energy evolution of the modal Fréchet distance at the end-effector. All the PCC models achieve a good match with the FE model. Note indeed that the Fréchet distance is always less than $6~[\si{\cm}]$. Furthermore, the accuracy improves significantly as more bodies are considered. For the $x$ component, the distance between the PCC and FE models decreases with the order of discretization, as expected. However, the fifth order PCC model does not always show the best match with the rigid robot along the $z$ direction. Nevertheless, the modal integral Fréchet distance in Fig.~\ref{fig:f int measure} shows that the five DoFs PCC model better fits the rigid motion overall. These results demonstrate that energy-frequency relation is not an exhaustive similarity measure in general. Moving to the coherence, see Fig.~\ref{fig:g measure}, the five DoFs model has good performance only for the first mode. The same conclusions can be drawn also for the modal integral coherence in Fig.~\ref{fig:g int measure}. Based on the overall performance, we conclude that the PCC robot with four DoFs has modal evolutions that overlap with those of the rigid arm at the tip more closely. 

\section{Conclusions and Future Work}
\label{sec:conclusions}
This paper presents the first investigation of eigenmanifolds for control-oriented reduced-order modeling of continuum soft robots. To improve computational efficiency, we present a new continuation algorithm for mode computation with an energy-based term. We propose a systematic approach to compare the eigenmanifolds of dynamic models resulting from different discretization hypotheses and number of degrees of freedom (DoFs). To this end, we project the eigenmanifolds onto a task space of fixed dimension and introduce therein different similarity measures. The proposed method is used to compare piecewise constant curvature (PCC) models of increasingly finer discretization with a finite element (FE) model. The results show that the task space modal evolutions of the PCC and FE models become similar when the order of the PCC discretization increases. 

Future work will explore the use of the proposed similarity measures to obtain novel control-oriented models with a reduced number of DoFs.


\bibliographystyle{IEEEtran}
\bibliography{bibliography}
\end{document}